\documentclass[11pt]{article}

\usepackage{acl}

\usepackage{times}
\usepackage{latexsym}

\usepackage[T1]{fontenc}

\usepackage[utf8]{inputenc}

\usepackage{booktabs}
\usepackage[table]{xcolor}
\usepackage{tabularx}
\usepackage{multirow}
\usepackage{adjustbox}
\usepackage{xcolor}
\usepackage{graphicx}
\usepackage{rotating}
\usepackage{longtable}
\usepackage{enumitem}
\usepackage{alltt}
\usepackage{hyphenat}
\usepackage{amsmath,amssymb}
\usepackage{placeins}
\usepackage{threeparttable}
\usepackage{float}
\usepackage{makecell}
\usepackage{hyperref}
\usepackage{array}

\title{Style over Story: Measuring LLM Narrative Preferences via Structured Selection}

\author{
  Donghoon Jung\thanks{Equal contribution.} \quad
  Jiwoo Choi\footnotemark[1] \quad
  Songeun Chae\footnotemark[1] \quad
  Seohyon Jung\thanks{Corresponding author.} \\
  School of Digital Humanities and Computational Social Sciences, KAIST, South Korea \\
  \texttt{\{donghoon.jung, jwchoi0515, songeun, seohyon.jung\}@kaist.ac.kr}
}

\date{}

\usepackage{graphicx}
\usepackage{caption} 
\usepackage{subcaption}
\usepackage{placeins}  
\graphicspath{{figures/}}
\begin{document}
\maketitle
\begin{abstract}

We introduce a constraint-selection-based experiment design for measuring narrative preferences of Large Language Models (LLMs). This design offers an interpretable lens on LLMs' narrative selection behavior. We developed a library of 200 narratology-grounded constraints and prompted selections from six LLMs under three different instruction types: basic, quality-focused, and creativity-focused. Findings demonstrate that models consistently prioritize \emph{Style} over narrative content elements like \emph{Event}, \emph{Character}, and \emph{Setting}. Style preferences remain stable across models and instruction types, whereas content elements show cross-model divergence and instructional sensitivity. These results suggest that LLMs have latent narrative preferences, which should inform how the NLP community evaluates and deploys models in creative domains. 
 
\end{abstract}

\section{Introduction}
Understanding the latent narrative preferences of large language models (LLMs) becomes more important as novelists are beginning to explore the use of LLMs in the writing process \citep{hern2023scifi, lewsey2025cambridge, robertson2025bookbub, authorsguild2023survey}. Prior works suggest that LLM use can reduce diversity in narrative plots \citep{xu2025echoes}, collective creativity \citep{doshi2024generative}, and individual writing styles \citep{osullivan2025stylometric}. These narrative and stylistic outcomes are significant because they may reflect unexplored preferences and biases of LLMs.

A growing line of research shows that LLM systems encode preference-related signatures, including political preferences \citep{rozado2024political}, personality traits \citep{serapiogarcia2025personality}, and value correlations \citep{rozen2025llms}. We extend this agenda to the narrative domain and reveal narrative preferences by eliciting choices among narrative constraints when alternatives are explicitly specified. Although some studies report discontinuities between self-reports and downstream behavior \citep{han2025personality, xu2025large}, others suggest that preference signals can remain coherent under within-domain measurement and can align with analyses of produced text \citep{goyanes2025personality, rozado2025measuring}.

Building on preference-profiling approaches, we treat selection behavior as a window into underlying priorities that output analysis cannot directly assess. Existing scholarship on LLMs and narrative has focused on analyzing generated outputs \citep{Chakrabarty2024Art, GomezRodriguez2023Confederacy, huot2025, ismayilzada2025, xie2024, Yang2022Re3}. Although these studies provide important evidence about LLM-generated narratives, including plot coherence or linguistic complexity, they cannot directly characterize latent narrative preferences. To complement output-centered approaches that require subjective quality judgments and conflate preference with ability, we propose structured selections from a set of narrative constraints. By asking models to select rather than generate, we isolate preference from production capacity and create conditions for controlled comparison across models.

We introduce a narratology-informed selection task in which models choose from a library of 200 narrative constraints, enabling quantitative comparison across LLMs. Our contributions are fourfold: (1) a constraint-selection methodology for measuring LLMs' narrative preference as a controlled alternative to output-centered evaluations, (2) a reusable library of narrative constraints with axis annotations, (3) evidence of systematic element-level preferences and their stability across models and instruction types, and (4) axis-level analysis that makes latent preferences interpretable beyond category level, revealing systematic shifts under a creativity-oriented instruction type.

\section{Related Work}
\subsection{Measuring Preferences in LLMs}

Recent studies have profiled LLM systems by eliciting preferences through structured instruments. \citet{rozado2024political} measures political preferences across LLMs, \citet{serapiogarcia2025personality} administer personality inventories with attention to psychometric quality, \citet{zheng2025lmlpa} measure personality through open-ended responses with an AI rater to support reliability and validity checks, and \citet{rozen2025llms} examine value rankings and their consistency across probes. Collectively, these studies argue that making latent preference-related profiles of LLM systems legible is important; \citet{rozado2024political}, for instance, is motivated by the concern that model parameters may encode crystallized assumptions with substantial societal influence as LLMs become widespread information sources.

Despite concerns about self-report validity \citep{han2025personality, xu2025large}, several studies identify conditions under which preference signals remain consistent. \citet{rozado2025measuring} reports convergence between standardized political diagnostics and text-based analyses, \citet{rozen2025llms} show coherent value structures across probes when values are anchored through instructions, and \citet{goyanes2025personality} finds weak links between broad personality measures and political attitudes in LLMs, but much stronger associations with domain-specific political background variables. \citet{wang2025comparative} argues that LLM responses are not fixed traits but depend on input context. These findings suggest that within-domain structured elicitation like our constraint-selection design can yield reliable results for estimating narrative preferences.

\subsection{Output-Centered Evaluation of LLM-Generated Narratives}
LLM-based narrative research has analyzed generated outputs and evaluated narrative quality. LLMs often produce locally fluent text \citep{Yang2022Re3} and linguistically complex prose \citep{ismayilzada2025}, and can score competitively on technical dimensions such as structure and readability \citep{GomezRodriguez2023Confederacy}. However, multiple studies report gaps in creativity-related dimensions, including originality and creative imagination \citep{GomezRodriguez2023Confederacy}, rhetorical complexity and nuanced character development \citep{Chakrabarty2024Art}, or novelty, surprise, and thematic diversity \citep{ismayilzada2025}. Long-form narrative generation further reveals difficulties with long-range plot coherence and premise relevance \citep{Yang2022Re3}, character development and meeting length constraints \citep{huot2025}, as well as suspense \citep{xie2024}. 

As noted above, a second set of findings concerns diversity. LLM-generated stories can show reduced plot diversity, including recurring “echoed” idiosyncratic narrative elements across outputs and models \citep{xu2025echoes} or reduced collective diversity \citep{doshi2024generative}. Stylometric evidence likewise suggests that LLM outputs exhibit notable stylistic uniformity and internal consistency unlike human writing \citep{osullivan2025stylometric}. These output-centered results highlight broad strengths and failure modes, but they cannot determine whether they stem from differences in model capabilities, training data biases, or systematic underlying preferences. Our design addresses this directly.

\subsection{Prompt Sensitivity and Measurement Robustness}
A central methodological concern for preference measurement is prompt sensitivity, where small changes in phrasing or context can substantially alter model responses \citep{lu-etal-2022-fantastically, zhuo-etal-2024-prosa}. Related work also documents order effects, showing that models can be sensitive to the position and ordering of inputs, which can distort apparent preferences in structured tasks \citep{liu-etal-2024-lost, pezeshkpour-hruschka-2024-large, shi2025judging}. \citet{shu-etal-2024-dont} further show that self-report-based personality measurements can shift under spurious prompt changes, raising concerns about the robustness of commonly used elicitation setups. 

We address these concerns by evaluating narrative preferences under multiple instruction types and by randomizing the presentation of the 200 narrative constraints across runs, reducing the influence of fixed ordering and presentation artifacts. We also report results across five experimental conditions with varied prompt framings, which allows us to examine preference stability across these conditions.

\section{Methodology}
We introduce a library of theory-grounded, structured narrative constraints to quantify LLM selection behavior as a proxy for latent narrative preferences, and use varied instruction types and task conditions to test the stability of these preferences across conditions.

For clarity, we adopt the following terminology. \emph{Narrative preference} denotes a systematically revealed priority structure: the ordering that emerges in constraint selection behavior under a controlled planning task with an explicitly framed prompt. \emph{Tendency} refers to the aggregate regularity and robustness of these selection patterns across runs and conditions. \emph{Selection behavior} denotes the specific act of selecting constraints; subsequent uses of "behavior" without further qualification refer to selection behavior in this task, not downstream generation behavior.

\subsection{Narrative Constraint Design}

\begin{figure}[H]
  \centering
  \includegraphics[width=\columnwidth]{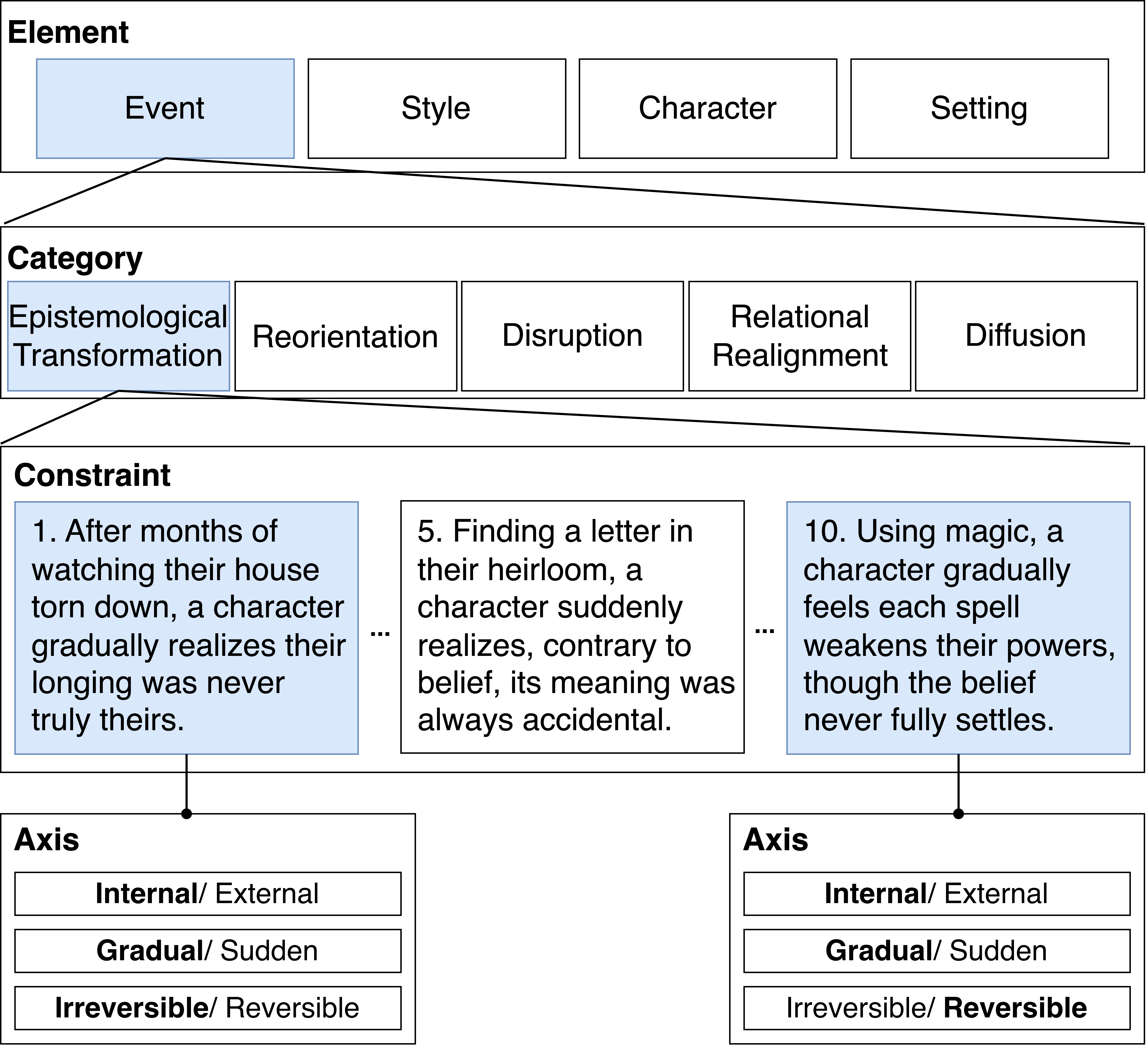}
  \caption{Structure of the Narrative Constraints Library. The library consists of four elements, with five categories per element and ten constraints per category. Each constraint is annotated using 1--3 axes that characterize them.}
  \label{fig:process}
\end{figure}

Narrative theories conceptualize narrative as a structured system built from distinct elements \citep{piper2021}. Based on classical and contemporary narratology, we constructed a library of 200 constraints systematically distributed across four core narrative elements: \emph{Event} involves plot dynamics and temporal structure \citep{genette1980, gius2022}; \emph{Style} reflects choices in voice, tone, and narration \citep{genette1980, phelan2009}; \emph{Character} concerns agency, roles, and interiority \citep{bal1997, piper2023}; and \emph{Setting} concerns spatial and contextual grounding for narrative interpretation \citep{ryan2015, ryan2016}. Each element is subdivided into five theoretically grounded categories that contain 10 constraints.

Categories operationalize core dimensions of each element, such as the change types for \emph{Event}, narrational and stylistic choices for \emph{Style}, agency and social positioning for \emph{Character}, and spatiotemporal and cultural scaffolding for \emph{Setting}. Each constraint is annotated with 1--3 axes to capture interpretable attribute variation within categories. These annotations are not shown to models and are used only for analysis.

To minimize surface-level selection bias, we standardize constraints to 15--20 words, parallel grammatical form, and matched conceptual granularity within categories. The full constraint list and axis annotations appear in \autoref{app:constraints}.

\subsection{Experiment Design}

\paragraph{Overview.}
We compare constraint selections across five task conditions grouped into three experiments. A \emph{run} is a single selection-and-justification response to a randomized constraint list under a fixed (model, instruction type, condition); decoding settings are held constant where available (see \autoref{app:models_decoding}).

\begin{figure}[H]
  \centering
  \includegraphics[width=\columnwidth]{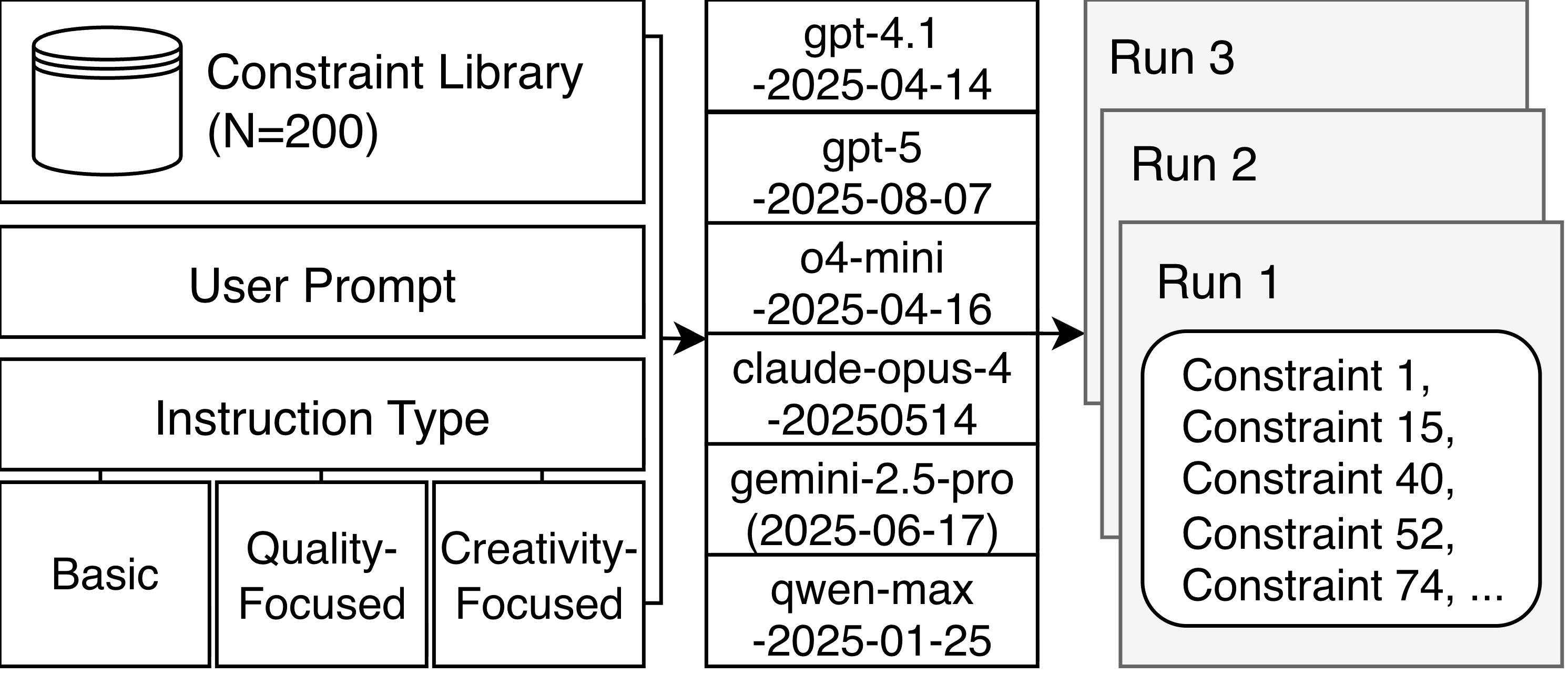}
  \caption{Experimental setup: models select constraints under different instruction types and task conditions with randomized candidate order.}
  \label{fig:exp}
\end{figure}

\paragraph{User prompt.}
User prompts standardize the task: depending on the condition, models select constraints either single-element (with a free or fixed budget) or from the pooled list (with or without a fixed budget), justify each selection, and provide a compatibility analysis. Constraint order is randomized each run to mitigate order effects. The user prompt for the baseline condition appears in \autoref{app:userprompt}.

\paragraph{Instruction type.}
We operationalize three instruction types as elicitation conditions that assign distinct functional framings to the LLMs:
\begin{itemize}
    \item \textbf{Basic:} established as a default instruction-following mode without particular leaning.
    \item \textbf{Quality-focused:} configured to prioritize technical excellence, including structural, plot-level cohesion and thematic integration.
    \item \textbf{Creativity-focused:} optimized toward non-conventional narrative patterns by weighting experimental and innovative selections over standard ones.
\end{itemize}
For brevity, we refer to these instruction types as Basic, Quality, and Creativity throughout the paper. Full instruction types appear in \autoref{app:system}.

\paragraph{Design and size.}
We cross six commercial LLMs, three instruction types (Basic, Quality-focused, Creativity-focused), and five task conditions. Prioritizing external validity over architectural ablations, we evaluate models that practicing novelists actually use: the two most commonly used platforms, GPT and Claude \citep{robertson2025bookbub}; two further GPT-family models (o4mini and gpt5 alongside gpt4.1) to probe within-family variation; and Gemini and Qwen as major models from diverse regional ecosystems for broader cross-provider coverage at comparable scale. Our findings thus target real-world impacts on literary practice rather than causal attribution to factors such as scale, training stage, or architecture. 

Stage~1 runs 30 independent replications per cell; for the element-wise conditions (1--1, 1--2), each replication yields four element-specific runs (event/style/character/setting). Stage~2 adds 160 replications per cell for the baseline condition (Experiment~2--2) to stabilize element- and category-level estimates. Each Stage~1 replication therefore produces 11 runs (4 each from conditions 1--1 and 1--2, plus one each from 2--1, 2--2, and 3). In total:
\[
N \;=\; M \times 3 \times \bigl(11 \times 30 + 160\bigr) \;=\; 8{,}820
\]
(Stage~1: 5{,}940; Stage~2: 2{,}880; \(M{=}6\)). The RR-based power analysis motivating \(R{=}160\) is reported in {\renewcommand{\subsectionautorefname}{Appendix}\autoref{app:stats_outcome}}.

\paragraph{Materials.}
All runs draw from the same library of 200 narrative constraints. Element labels are visible in element-wise and labeled pooled tasks (1--1, 1--2, 3) and hidden in pooled unlabeled tasks (2--1, 2--2); axis annotations are never shown to models.

\paragraph{Task conditions.}
We evaluate five conditions: single-element labeled selection with free budget (1--1) or fixed budget ($K{=}5$; 1--2), pooled unlabeled selection with free budget (2--1) or fixed budget ($K{=}20$; 2--2), and element-blocked labeled selection with quotas ($K{=}20$, $k{=}5$ per element; 3). Each run outputs selected constraints with per-constraint justifications and a compatibility analysis (element coverage is inferred for pooled unlabeled tasks).

\paragraph{Randomization and replication.}
Each run uses a fresh random permutation of the candidate list(s) and an isolated session state. Within a cell, replications vary only by permutation and stochastic decoding; prompts and candidate sets are otherwise identical.

\paragraph{Baseline condition.}
We use Experiment~2--2 (pooled, unlabeled, \(K{=}20\)) as the primary reference condition; the rationale is established via condition contrasts in Results.

\subsection{Outcomes and Analysis}

\paragraph{Scope.}
Experiment~2--2 (pooled, unlabeled, \(K{=}20\)) is the baseline. Cross-condition comparisons use Stage~1 replications; fine-grained analyses of the baseline use Stage~2 (or pooled Stage~1+2 where stated). Let \(y\) denote selections and \(s=y/K\) the within-run selection share; supply is \(n\) available candidates (see ~\autoref{app:stats} for full definitions).

\paragraph{Outcomes.}
We analyze (i) condition contrasts in selection shares with supply controls, (ii) per-run element and category compositions, and (iii) axis enrichment measured as observed-to-expected ratios. In our data, all runs have \(K_u \ge 1\) (no zero-selection responses), so selection shares \(s_{uc}=y_{uc}/K_u\) are well-defined for all runs.

\paragraph{Models and inference.}
We estimate condition contrasts using $K$-weighted WLS on selection shares with run-clustered SEs. Element- and category-level compositions are modeled with run-clustered Poisson GEEs using an exposure offset, and axis enrichment is assessed via stratified permutation tests; full specifications, offsets, and multiple-testing procedures are reported in {\renewcommand{\subsectionautorefname}{Appendix}\autoref{app:stats_outcome}}, {\renewcommand{\subsectionautorefname}{Appendix}\autoref{app:stats_gee}}, {\renewcommand{\subsectionautorefname}{Appendix}\autoref{app:stats_axis}}.

\section{Results}
In this section, we characterize LLM selection behavior across hierarchical levels, moving from broad narrative elements and mid-level categories to attribute axes. We report element- and category-level selection patterns and identify statistically salient constraints to reveal structural patterns underlying latent narrative preferences.

\subsection{Comparison of Experimental Setups}
We begin by evaluating the varying experimental setups to establish the baseline condition that grounds all subsequent analyses.

\paragraph{Outcome \& modeling.}
We analyze category-level selection shares and estimate condition contrasts under supply adjustment. Full outcome definitions, covariate adjustment, and contrast coding are reported in {\renewcommand{\subsectionautorefname}{Appendix}\autoref{app:stats_outcome}}.

\subsubsection{Selecting the Baseline Condition through Condition Contrasts}

As shown in \autoref{tab:condition_contrasts}, the condition contrasts identify a clear baseline. The pooled, unlabeled, fixed-budget setup (Experiment~2--2) leaves models closest to their native preference structure. Removing element labels limits priming. Foregoing element-wise constraints also avoids artificial portfolios that otherwise push stylistic mimicry or spatial specifics at the expense of abstract transformation or affect control. Fixing the selection budget ($K{=}20$) also yields more stable behavior across models and instruction types and simplifies inference, with transparent fixed effects and supply adjustments. Consistent with our narratology-informed aim to observe process-level authorial choices rather than engineer them, we adopt 2--2 as an interpretable reference for all cross\hyp{}model and cross\hyp{}prompt comparisons.

\paragraph{Model adequacy.}
Adequacy and robustness checks for the run--clustered Poisson GEE specifications (dispersion, working correlation) are reported in {\renewcommand{\subsectionautorefname}{Appendix}\autoref{app:stats_adequacy}}.

\begin{table}[H]
\centering
\footnotesize
\setlength{\tabcolsep}{8pt}
\renewcommand{\arraystretch}{1.1}
\begin{tabular}{l l}
\toprule
\textbf{Contrast} & \textbf{Largest shifts (pp)} \\
\midrule
1--2 vs.~1--1 & \emph{Epistemological Transformation} $+5.65$ \\
              & \emph{Embodied Difference} $-3.20$ \\
\midrule
2--2 vs.~2--1 & \emph{Cultural context} $+1.47$ \\
              & \emph{Narrative perspective} $-2.93$ \\
\midrule
3 vs.~1--2    & \emph{Write like X} $+14.97$ \\
              & \emph{Epistemological Transformation} $-20.63$ \\
\midrule
3 vs.~2--2    & \emph{Motive} $+3.48$ \\
              & \emph{Tone \& Mood} $-3.80$ \\
\bottomrule
\end{tabular}
\caption{Condition contrasts in covariate-adjusted category shares (pp) estimated by $K$-weighted WLS. For each contrast, entries report the largest positive and negative category shifts; positive values indicate higher selection under the first-listed condition.}
\label{tab:condition_contrasts}
\end{table}

\subsection{Element-Level Selection Patterns: Style Over Story}

With the baseline established, we next examine element-level selection patterns to see how models allocate preferences across \emph{Style}, \emph{Character}, \emph{Event}, and \emph{Setting}.

\paragraph{Model \& inference.}
We model run--element counts using Poisson GEEs clustered by run and report Poisson rate ratios (RRs) and selected pairwise contrasts. Full specifications and reporting conventions are in {\renewcommand{\subsubsectionautorefname}{Appendix}\autoref{app:stats_gee_element}} and {\renewcommand{\subsubsectionautorefname}{Appendix}\autoref{app:stats_inference_element}}.

\subsubsection{The Primacy of Style: Global Preference Structure}

LLMs showed a clear preference structure across elements (\autoref{tab:element_main}). Constraints in \emph{Style} were chosen most frequently, \emph{Setting} constraints were also selected more often than the baseline, and \emph{Character} did not differ from \emph{Event}. This pattern suggests that models place greater weight on expressive form and stylistic modulation (tone, register, voice) than story-driving elements such as \emph{Event} and \emph{Character}.

\begin{table}[H]
\centering
\small
\setlength{\tabcolsep}{9pt}
\begin{tabular}{lcc}
\toprule
\textbf{Element} & \textbf{RR [95\% CI]} & \textbf{$p$} \\
\midrule
\emph{Event} (baseline) & 1.00 [1.00, 1.00]      & ---     \\
\emph{Style}            & 1.78 [1.74, 1.82] & $<.001$ \\
\emph{Character}        & 0.98 [0.96, 1.01] & $.160$ \\
\emph{Setting}          & 1.28 [1.25, 1.31] & $<.001$ \\
\bottomrule
\end{tabular}
\caption{\label{tab:element_main}
Element-level selection rate ratios (RRs) vs.\ \emph{Event} (baseline) from a Poisson GEE clustered by run with offset $\log K$ (\(N=2{,}880\) runs). RR\(>1\) indicates more frequent selection than \emph{Event}.}
\end{table}

\subsubsection{Cross-Model Stability in Element Preferences}

Element-level preferences are broadly similar across models, with relatively small differences in selection behavior. Most systems allocate selections in broadly similar proportions across \emph{Style}, \emph{Character}, \emph{Event}, and \emph{Setting}. An exception is gpt4.1, which is more strongly tilted toward \emph{Style} and correspondingly less oriented toward \emph{Event}, \emph{Character}, and \emph{Setting} in the strongest contrasts. \autoref{tab:element_model_top3} makes this asymmetry visible across all four elements: gpt4.1 sits at one end of every largest pairwise contrast, as the most \emph{Style}-leaning model and simultaneously the model least oriented toward \emph{Event}, \emph{Character}, and \emph{Setting}. In this sense, gpt4.1 functions as an amplified instance of the broader tendency to prioritize expressive control. Overall, cross-model variation in element preferences remains moderate once gpt4.1 is set aside, with the remaining models clustering more closely around the global element ordering.

\begin{table}[H]
\centering
\small
\setlength{\tabcolsep}{6pt}
\renewcommand{\arraystretch}{1.12}
\begin{tabular}{c l c c}
\toprule
\textbf{Element} & \textbf{Contrast} & \(\boldsymbol{\Delta}\) \textbf{(\%)} & \(\boldsymbol{q}\) \\
\midrule
          & gpt5 $>$ gpt4.1         & +28 & $<.001$ \\
\emph{Event}     & gpt4.1 $<$ gemini        & -21 & $<.001$ \\
          & qwen $<$ gpt5            & -19 & $<.001$ \\
\midrule
          & gpt4.1 $>$ gemini       & +64 & $<.001$ \\
\emph{Style}     & gpt5 $<$ gpt4.1         & -35 & $<.001$ \\
          & o4mini $<$ gpt4.1       & -32 & $<.001$ \\
\midrule
          & qwen $>$ gpt4.1         & +32 & $<.001$ \\
\emph{Character} & gpt4.1 $<$ claude       & -22 & $<.001$ \\
          & gpt5 $>$ gpt4.1         & +27 & $<.001$ \\
\midrule
          & o4mini $>$ gpt4.1       & +71 & $<.001$ \\
\emph{Setting}   & gpt5 $>$ gpt4.1         & +51 & $<.001$ \\
          & qwen $>$ gpt4.1         & +43 & $<.001$ \\
\bottomrule
\end{tabular}
\caption{For each element, the three largest BH--FDR-significant pairwise model contrasts ($q\le 0.05$). Effects are reported as $\Delta=(\mathrm{RR}-1)\times 100$ for the first-listed model relative to the second.}
\label{tab:element_model_top3}
\end{table}

\subsubsection{\emph{Style} Stability with Creativity-Driven Shifts}

Overall element-level preferences remain largely stable across instruction types (\autoref{tab:prompt_element_contrasts}). The most noticeable shift occurs under the Creativity instruction type: relative to Basic and Quality, Creativity reweights selections away from \emph{Event} and \emph{Character} and uniquely toward \emph{Setting}. These shifts are statistically detectable despite modest effect sizes. Quality and Basic are effectively indistinguishable for Event and Character. \emph{Style} is the only element that shows no meaningful differences across instruction types.

\begin{table}[H]
\centering
\small
\setlength{\tabcolsep}{5pt}
\renewcommand{\arraystretch}{1.2}
\begin{adjustbox}{max width=\columnwidth}
\begin{tabular}{l l c c}
\toprule
\textbf{Element} & \textbf{Contrast} & \(\boldsymbol{\Delta}\) \textbf{(\%)} & \(\boldsymbol{q}\) \\
\midrule
\emph{Event}
  & Creativity $<$ Basic   & -19 & $<.001$ \\
  & Quality $>$ Creativity & +22 & $<.001$ \\
\midrule
\emph{Character}
  & Creativity $<$ Basic   & -29 & $<.001$ \\
  & Quality $>$ Creativity & +43 & $<.001$ \\
\midrule
\emph{Setting}
  & Creativity $>$ Basic   & +36 & $<.001$ \\
  & Quality $<$ Creativity & -30 & $<.001$ \\
\bottomrule
\end{tabular}
\end{adjustbox}
\caption{Instruction-type pairwise contrasts by element (BH--FDR, $q\le .05$). Significant differences are observed for \emph{Event}, \emph{Character}, and \emph{Setting}; none are observed for \emph{Style}. Effects are $\Delta=(\mathrm{RR}-1)\times 100$ for the first-listed prompt vs.\ the second.}
\label{tab:prompt_element_contrasts}
\end{table}

\subsection{Category-Level Selection Patterns: Common Style, Functional Divergence in Story}

After establishing differences across elements, we probe category-level patterns to uncover finer distinctions within each narrative dimension.

\paragraph{Modeling and inference.}
Within each element, we fit run-clustered Poisson GEEs to run-level category counts and report category rate ratios relative to a within-element reference category, along with within-category pairwise contrasts.
Inference uses robust (sandwich) SEs with Wald tests and BH--FDR correction; full specifications, offsets, and reporting rules are in {\renewcommand{\subsubsectionautorefname}{Appendix}\autoref{app:stats_gee_category}}, {\renewcommand{\subsubsectionautorefname}{Appendix}\autoref{app:stats_inference_category}}.

\subsubsection{The Shared Canon of \emph{Style}: Dominance of \emph{Tone \& Mood}}

\newcommand{\hi}{\(\uparrow\)}
\newcommand{\lo}{\(\downarrow\)}

\begin{table}[H]
\centering
\small
\setlength{\tabcolsep}{4pt}
\renewcommand{\arraystretch}{1.15}
\begin{adjustbox}{max width=\columnwidth}
\begin{tabular}{@{}m{.36\columnwidth} m{.64\columnwidth}@{}}
\toprule
\textbf{Element (baseline)} & \textbf{Selection-rate differences ($\Delta$\%)} \\
\midrule
\emph{Event} \emph{(Diffusion)} &
\textbf{\hi} \emph{Epistemological Transformation} \(+115\%\) \\
 &
\textbf{\hi} \emph{Reorientation} \(+50\%\) \\
\midrule
\emph{Style} \emph{(Narrative} &
\textbf{\hi} \emph{Tone \& Mood} \(+127\%\) \\
\emph{ perspective)} &
\textbf{\lo} \emph{Write like X} \(-65\%\) \\
\midrule
\emph{Character} &
\textbf{\hi} \emph{Motive} \(+109\%\) \\
\emph{(Cultural Identity)} &
\\
\midrule
\emph{Setting} &
\textbf{\hi} \emph{Temporal setting} \(+57\%\) \\
\emph{(Cultural context)} &
\\
\bottomrule
\end{tabular}
\end{adjustbox}
\caption{\label{tab:category_main}
Large category shifts vs.\ within-element baseline categories in Experiment~2--2, reported as $\Delta\%=(\mathrm{RR}-1)\times 100$.
We show categories with $p<.05$ and $|\Delta|\ge 50\%$.
A complete table of RRs appears in \autoref{app:cat_main_rr}.}
\end{table}

At the category level, within \emph{Style}, the \emph{Tone \& Mood} category is most prominent, while \emph{Write like X} receives considerably less preference. \emph{Tone \& Mood} captures constraints that set the narrative's affective atmosphere and perceptual texture (see \autoref{app:constraints}). This within-style imbalance suggests limited appetite for explicit authorial mimicry (\emph{Write like X}). Across the other elements, \emph{Event} shows elevated \emph{Epistemological Transformation} and \emph{Reorientation}, \emph{Character} emphasizes \emph{Motive}, and \emph{Setting} prioritizes \emph{Temporal setting}, each measured against a within-element baseline category (see \autoref{tab:category_main}).

\subsubsection{Cross-Model Patterns: \emph{Style} Convergence with Authorial Outliers}

\begin{table}[H]
\centering
\small
\setlength{\tabcolsep}{5pt}
\renewcommand{\arraystretch}{1.2}
\begin{adjustbox}{max width=\columnwidth}
\begin{tabular}{l c c}
\toprule
\textbf{Category} & \textbf{Contrast} & \(\boldsymbol{\Delta}\) \textbf{(\%)} \\
\midrule
\multicolumn{3}{l}{\emph{\textbf{Event}}} \\
\emph{Diffusion}                      & o4mini $<$ gpt5     & -41  \\
\emph{Disruption}                     & o4mini $>$ gemini   & +209 \\
\emph{Epistemological Transformation} & gpt4.1 $<$ gemini   & -32  \\
\emph{Relational Realignment}         & qwen $>$ gpt5       & +237 \\
\emph{Reorientation}                  & o4mini $<$ gpt4.1   & -36  \\
\midrule
\multicolumn{3}{l}{\emph{\textbf{Character}}} \\
\emph{Cultural Identity}              & gpt4.1 $>$ claude   & +88  \\
\emph{Embodied Difference}            & qwen $>$ gpt5       & +135 \\
\emph{Motive}                         & o4mini $>$ gpt4.1   & +47  \\
\emph{Relational Identity}            & gpt5 $<$ gemini     & -40  \\
\emph{Social Status}                  & gpt4.1 $>$ gemini   & +67  \\
\midrule
\multicolumn{3}{l}{\emph{\textbf{Setting}}} \\
\emph{Cultural context}               & o4mini $>$ gpt4.1   & +157 \\
\emph{Macro spatial setting}          & gpt4.1 $>$ gemini   & +99  \\
\emph{Micro spatial setting}          & o4mini $<$ gpt5     & -47  \\
\emph{Socio\hyp{}political order}     & o4mini $>$ claude   & +116 \\
\emph{Temporal setting}               & qwen $>$ gemini     & +42  \\
\midrule
\multicolumn{3}{l}{\emph{\textbf{Style}}} \\
\emph{Narrative perspective}          & qwen $>$ o4mini     & +67  \\
\emph{Temporal Structure}             & o4mini $<$ gpt4.1   & -15 \\
\emph{Syntax \& Sentence Structure}   & gpt4.1 $<$ claude   & -52  \\
\emph{Tone \& Mood}                   & qwen $<$ o4mini     & -55  \\
\emph{Write like X}                   & qwen $>$ gemini     & +642 \\
\bottomrule
\end{tabular}
\end{adjustbox}
\caption{Within each category, we report the single largest (by $|\Delta\%|$) BH--FDR-significant pairwise model contrast ($q \le 0.05$), where $\Delta\%=(\mathrm{RR}-1)\times 100$. Positive (negative) $\Delta\%$ indicates higher (lower) selection for the first-listed model. Values are rounded.}
\label{tab:model_category_top1}
\end{table}

Cross-model variability appears across all four elements, but within \emph{Style} it concentrates on authorial-mimicry categories. \emph{Write like X} shows the largest contrast by a wide margin, driven by individual outliers: gpt5 never selected \emph{Write like X}, while qwen over-selected it at a level far exceeding any other cross-model contrast in the study. Other \emph{Style} categories show cross-model differences as well, but on a much smaller scale than \emph{Write like X} (\autoref{tab:model_category_top1}).

\subsubsection{Instructional Stability in \emph{Style} and \emph{Setting}, Sensitivity in \emph{Event} and \emph{Character}}

At the category level, instructional differences remain limited and are primarily driven by the Creativity instruction type. \emph{Style} shows no meaningful differences across instruction types within its subcategories, indicating that prompt effects are negligible for this dimension. \emph{Setting} shows at most a limited separation, with differences that remain modest in magnitude. In contrast, instructional sensitivity concentrates in \emph{Event} and \emph{Character}. Within \emph{Event}, Creativity decreases emphasis on \emph{Diffusion} and \emph{Relational Realignment}, which Quality strongly favors. Within \emph{Character}, Creativity favors \emph{Embodied Difference} while down-weighting \emph{Social Status}; Quality shows the opposite pattern.

\begin{table}[H]
\centering
\small
\setlength{\tabcolsep}{5pt}
\renewcommand{\arraystretch}{1.2}
\begin{adjustbox}{max width=\columnwidth}
\begin{tabular}{l c c}
\toprule
\textbf{Category} & \textbf{Contrast} & \(\boldsymbol{\Delta}\) \textbf{(\%)} \\
\midrule
\multicolumn{3}{l}{\emph{\textbf{Event}}} \\
\emph{Diffusion}                    & Quality $>$ Creativity    & +71  \\
\emph{Relational Realignment}       & Quality $>$ Creativity    & +110 \\
       & Creativity $<$ Basic      & -52 \\
\midrule
\multicolumn{3}{l}{\emph{\textbf{Character}}} \\
\emph{Embodied Difference}          & Creativity $>$ Basic      & +90  \\
          & Quality $<$ Creativity    & -53 \\
\emph{Social Status}                & Quality $>$ Creativity    & +139 \\
                & Creativity $<$ Basic      & -53 \\
\bottomrule
\end{tabular}
\end{adjustbox}
\caption{Instruction-type contrasts by category (Poisson GEE), reporting only BH--FDR significant results with $q<.001$ and $|\Delta\%|\ge 50$, where $\Delta\%=(\mathrm{RR}-1)\times 100$. Positive (negative) $\Delta\%$ indicates higher (lower) selection for the first-listed prompt. No \emph{Style} category survives BH--FDR; \emph{Setting} survives BH--FDR but not the $|\Delta\%|\ge 50$ threshold. Values are rounded.}

\label{tab:prompt_category_contrasts}
\end{table}

\subsection{Axis-Level Patterns: Surfacing Abstract Narrative Preferences}

With category-level patterns established, we next examine axis-level patterns to see how selection mass concentrates on abstract narrative attributes and how these concentrations shift by instruction type. We map chosen constraints onto curated axes and summarize how selection mass concentrates at this level. We first establish a global axis baseline by comparing observed selections to supply-adjusted expectations aggregated across runs, and then examine axis shifts by instruction type using directional over and under signals from the constraint-level tests (see \autoref{app:axis_top5_instruction_type}). We treat these axis summaries as descriptive guides rather than as an additional layer of statistical inference. We report model-stratified summaries in \autoref{app:axis_model_summaries}.

\paragraph{Method summary.}
Within Experiment~2--2, we compare observed selections to supply-adjusted expectations and aggregate over/under-selection signals at the axis level; details are in {\renewcommand{\subsectionautorefname}{Appendix}\autoref{app:stats_axis}}.

\subsubsection{Presentist Anchoring with Selective Departures}

\autoref{tab:axis_global_baseline} summarizes a global axis baseline by showing which narrative attributes receive disproportionate selection mass after supply adjustment. The dominant pattern is presentist anchoring. Models over-select everyday contemporary scaffolds such as \emph{Urban Built Environments} and \emph{The Fully Connected Now}, while down-weighting distant historical frames such as \emph{Age of Origins}. Alongside these anchors, models also over-select departures in space and time, including \emph{Dreamlike or Surreal Chambers} and \emph{The Broken Sequence}, while \emph{Mythic or Enchanted Structures} remains under-selected. The baseline also tilts toward \emph{Positive} reorientation and away from \emph{Unquestioned Precedent}.

\begin{table}[H]
\centering
\small
\setlength{\tabcolsep}{6pt}
\renewcommand{\arraystretch}{1.2}
\begin{tabularx}{\columnwidth}{@{}l X@{}}
\toprule
\textbf{Direction} & \textbf{Axis (Element, Category) (Obs/Exp $\times$)} \\
\midrule
\textbf{Over} &
\emph{\textbf{Urban Built Environments}} (\textit{Setting, Macro spatial setting}) (3.66),
\emph{\textbf{The Fully Connected Now}} (\textit{Setting, Temporal setting}) (2.74),
\emph{\textbf{Dreamlike or Surreal Chambers}} (\textit{Setting, Micro spatial setting}) (2.52),
\emph{\textbf{The Broken Sequence}} (\textit{Setting, Temporal setting}) (1.85),
\emph{\textbf{Positive}} (\textit{Event, Reorientation}) (2.06) \\
\midrule
\textbf{Under} &
\emph{\textbf{Age of Origins}} (\textit{Setting, Temporal setting}) (0.17),
\emph{\textbf{Mythic or Enchanted Structures}} (\textit{Setting, Micro spatial setting}) (0.33),
\emph{\textbf{Unquestioned Precedent}} (\textit{Setting, Cultural context}) (0.26) \\
\bottomrule
\end{tabularx}
\caption{\label{tab:axis_global_baseline}
Global axis baseline (supply-adjusted; Obs/Exp). We list only the axes explicitly discussed in the text. Axes are shown with their element and category for disambiguation. Values rounded. The full top-15 global axes table appears in \autoref{app:global_axis_enrichment}. We treat \emph{Connected} axis under reorientation category separately because it primarily indexes event-level linkage.}
\end{table}

\subsubsection{Shifts under Creativity away from Everyday Realism}

\autoref{tab:axis_prompt_commonUnion20_bycat} summarizes how instruction types redistribute significance-filtered constraint signals across narrative axes relative to the global baseline. Basic and Quality show greater enrichment on everyday realist scaffolds, including \emph{Urban Built Environments}, \emph{Domestic Interior Spaces}, and \emph{Transit Hubs}, and on the presentist temporal frame, \emph{The Fully Connected Now}. On these same axes, Creativity shifts selections away from everyday realist anchoring. A parallel contrast appears in perspective. Creativity is relatively enriched on \emph{Second person perspective}, whereas Basic and Quality show the opposite tendency. Overall, these contrasts suggest a structured shift under Creativity rather than isolated axis-by-axis variation.

\providecommand{\hi}{\ensuremath{\uparrow}}
\providecommand{\lo}{\ensuremath{\downarrow}}

\newcommand{\axec}[3]{\textbf{#1} (\emph{#2}, \emph{#3})}

\newcommand{\axblock}[1]{\begin{minipage}[t]{\linewidth}\raggedright #1\end{minipage}}

\begin{table}[H]
\centering
\small
\setlength{\tabcolsep}{4pt}
\renewcommand{\arraystretch}{1.4}
\begin{adjustbox}{max width=\columnwidth}
\begin{tabular}{@{}%
  >{\raggedright\arraybackslash}p{.34\columnwidth}%
  >{\raggedright\arraybackslash}p{.66\columnwidth}@{}}
\toprule
\textbf{Prompt} & \textbf{Axis (Element, Category)} \\
\midrule
Basic, Quality \lo; Creativity \hi &
\axblock{%
  \axec{\textit{Second}}{Style}{Narrative perspective}%
} \\
\midrule
Basic, Quality \hi; Creativity \lo &
\axblock{%
  \axec{\textit{Urban Built Environments}}{Setting}{Macro spatial setting}\\
  \axec{\textit{Domestic Interior Spaces}}{Setting}{Micro spatial setting}\\
  \axec{\textit{Transit Hubs}}{Setting}{Micro spatial setting}\\
  \axec{\textit{The Fully Connected Now}}{Setting}{\textit{Temporal setting}}%
} \\
\bottomrule
\end{tabular}
\end{adjustbox}
\caption{Axes common to all instruction types. For each prompt we take the union of the top~20 axes from over and under (ranked by enrichment), intersect across prompts (direction-agnostic), and drop axes with a uniform direction. Left column shows the per-prompt direction relative to the global baseline (\hi\,=\,over; \lo\,=\,under).}
\label{tab:axis_prompt_commonUnion20_bycat}
\end{table}

\section{Discussion}
The pronounced preference for \emph{Style} indicates that LLMs prioritize stylistic dimensions over elements that define the substantive content of a narrative.
Importantly, this preference remains largely stable across instruction types. Although the Creativity instruction reweights selections across \emph{Event}, \emph{Character}, and \emph{Setting}, \emph{Style} alone is preserved.
Prior work on LLM capabilities provides a useful framework for interpreting this pattern.
LLM competence has been characterized in terms of formal linguistic competence and functional linguistic competence, with strong performance in the former but persistent limitations in the latter \citep{mahowald2024dissociating}.
Relatedly, maintaining coherence over long narrative contexts remains challenging for language models \citep{Yang2022Re3}.
In this context, one plausible interpretation is that the observed preferences for stylistic dimensions over content-related narrative elements reflect asymmetries in LLMs’ linguistic capabilities.

To investigate whether capability drives the observed narrative preferences, we focused on gpt4.1, the model that showed the strongest tilt toward \emph{Style} in our main analysis, and prompted it to select 20 constraints out of 200 under two conditions: the Better-story condition, in which the model identified constraints that "would contribute most to high overall narrative quality," and the Capability condition, in which it identified constraints that "would contribute most to ease of execution" (full prompts in \autoref{app:userprompt-variants}). \emph{Style} remained the most selected element across both conditions (\autoref{app:ratio-gpt4-new}). If style dominance merely reflected capability asymmetries, we would expect \emph{Style} selections to be equal or higher under the Capability condition. Instead, we observed the opposite pattern: relative to the Better-story condition, the Capability condition showed lower \emph{Style} selection and higher \emph{Event} and \emph{Character} selections (\autoref{tab:prompt_pairwise_betterstory_capability}). This divergence indicates that stylistic preference is not reducible to what LLMs can reliably execute; rather, this pattern suggests that stylistic dimensions are treated as central to what constitutes a "better narrative."

\begin{table}[H]
\centering
\small
\setlength{\tabcolsep}{5pt}
\renewcommand{\arraystretch}{1.2}
\begin{adjustbox}{max width=\columnwidth}
\begin{tabular}{l l c c}
\toprule
\textbf{Element} & \textbf{Contrast} & \(\boldsymbol{\Delta}\) \textbf{(\%)} & \(\boldsymbol{q}\) \\
\midrule
\emph{Event}     & Capability $>$ Better-story & +22 & $<.001$ \\
\emph{Style}     & Capability $<$ Better-story & $-11$ & $<.001$ \\
\emph{Character} & Capability $>$ Better-story & +13 & $.023$ \\
\bottomrule
\end{tabular}
\end{adjustbox}
\caption{User-prompt pairwise contrasts by element for gpt4.1 under two additional user-prompt variants (Better-story and Capability), applied under the same \emph{Basic} instruction type, using the exposure offset $\log K$. Values are reported with BH--FDR adjusted $q$. Significant differences ($q\le .05$) are observed for \emph{Event}, \emph{Character}, and \emph{Style}; the \emph{Setting} contrast is not significant and is omitted. Effects are $\Delta=(\mathrm{RR}-1)\times 100$ for the first-listed prompt vs.\ the second. \emph{N runs} $=320$.}
\label{tab:prompt_pairwise_betterstory_capability}
\end{table}

From a narratological point of view, our findings point beyond capability to a more fundamental question about LLMs' narrative generation. While "style over story" might be interpreted as general prioritization of form over content, traditional narrative theory maintains that style is not a mere formal layer but the core literary device that establishes author's voice and creative authority. The within-\emph{Style} skew toward \emph{Tone \& Mood} over specific authorial styles highlights that LLMs favor generalized and recognizable stylistic expression over the kind of voice that is specifically attributable. This reflects a broader preference for the typical over the particular. That models under the Creativity instruction shifted toward unrealistic temporal and spatial Settings such as surreal or extraterrestrial settings rather than toward Event or Character configurations further implies a typified understanding of creativity. Surface-level "unusualness" gets adopted as a sign of creativity rather than genuine and unique innovation in plot and characterization. These trends deserve further examination because they have direct consequences in examining how narratives are generated and evaluated.

Understanding LLMs' narrative preferences should be seen not as capturing incidental tendencies but as a prerequisite for robust system design.
The observed tendencies suggest that achieving controllability and diversity in narrative generation requires explicit consideration of structural preferences, rather than relying solely on surface level prompt adjustments.
More cautiously, our results point to a potential source of bias in automatic evaluation, particularly when LLMs are used as judges \citep{zheng2023judging}.
The underlying preferences may systematically favor specific narrative forms or contents.
Such tendencies may not be fully reflected in generated outputs alone, as stable stylistic preferences can remain implicit in unconstrained generation.
Our analysis therefore provides a necessary complement to existing evaluation practices.
From a methodological perspective, the use of axis annotations enables more interpretable experimental designs, as it allows for post-hoc analyses of latent preferences in LLMs.

Future work should examine whether the preferences identified in this study generalize across a wider range of narrative-related tasks.
Evidence from other domains suggests that they may: \citet{rozado2025measuring} and \citet{goyanes2025personality} show that preferences exhibit systematic patterns across related tasks, while \citet{rozen2025llms} report value consistency under fixed instruction types.
Given that our experimental setting relies on explicitly framed story-planning instructions, the observed preferences in this study may extend to other narrative-related tasks.
A second direction involves a finer-grained examination of cross-model differences.
We observe category-level differences within each content-related element—\emph{Event}, \emph{Character}, and \emph{Setting}—suggesting that finer-grained investigation could reveal distinct narrative landscapes across models. Making such preferences legible advances interpretability in LLM-assisted creative narrative tasks.

\section*{Limitations}
This study is scoped to English-language constraints and our prompt design. Narrative preferences may differ across languages and under prompts that are substantially shorter or longer than ours. We also focus on proprietary commercial models, so our findings may not generalize to open-weight systems with different training, fine-tuning, and alignment regimes. The 200 narrative constraints were constructed by the authors drawing on narratology scholarship and relevant research background, with an emphasis on comprehensive coverage and within-category consistency. However, the library is not externally validated and is not exhaustive for capturing all dimensions of narrative design. At the interpretive level, our design elicits preferences through a selection task with a restricted set of constraints, and preferences might manifest differently under alternative framings. Finally, we treat LLM selection behavior as a signal of latent preferences, but how these preferences manifest in downstream narrative generation under comparable settings remains to be examined. Selection remains a proxy and may also reflect capability or alignment constraints, and axis annotations reflect our design choices. We do not make fairness or social-bias claims from identity-related constraints, and proprietary API models (including the specific snapshot versions used here) may change over time or become unavailable, limiting exact replication.

\section*{Ethical Considerations}
Our findings are specific to this English constraint library and elicitation setup and should not be overgeneralized as universal properties of LLMs across languages, prompts, or applications. Because our constraint library explicitly includes identity-related attributes, repeated reuse of these options can make particular framings more salient and can encourage simplified associations about social groups when taken out of context. In addition, the constraint list directly names individual authors in the “Write like X” items, which can be sensitive and may be interpreted as endorsing or encouraging imitation.

\section*{Acknowledgments}
This work was supported by the Korea Advanced Institute of Science and Technology (KAIST) under the project "Quantifying Creativity: Developing Metrics for Evaluating AI-Generated Narratives" (Project No. 11250011 and No. N10260075).

\bibliographystyle{acl_natbib}
\bibliography{references}

\clearpage   
\appendix

\onecolumn

\renewcommand{\sectionautorefname}{Appendix}
\renewcommand{\subsectionautorefname}{Appendix}
\renewcommand{\subsubsectionautorefname}{Appendix}

\section{Constraints}\label{app:constraints}
\vspace{-0.5em}

\subsection{Event constraints (n=50)}
\vspace{-1em}
\begin{table}[H]
\centering
\setlength{\tabcolsep}{1.3pt}
\renewcommand{\arraystretch}{1}
\fontsize{8.5pt}{10pt}\selectfont
\begin{tabularx}{\textwidth}{@{}p{0.3cm} >{\raggedright\arraybackslash}X >{\raggedleft\arraybackslash}p{1.2cm}@{\hspace{0.15cm}}}
\toprule
\textbf{\#} & \textbf{Constraint} & \textbf{Axes} \\
\midrule
\multicolumn{3}{l}{\textbf{\textit{Epistemological Transformation}}}\\
1  & After months of watching their house torn down, a character gradually realizes their longing was never truly theirs. & I·G·X \\
2  & After overhearing a conversation, a character suddenly understands with certainty their loved one has lived a secret life. & E·S·X \\
3  & As memories gradually return, a character becomes aware that something they believed might be pure invention. & I·G·R \\
4  & A friend’s sudden confession makes a character decide to seek truth even if it puts them in danger. & E·S·R \\
5  & Finding a letter in their heirloom, a character suddenly realizes, contrary to belief, its meaning was always accidental. & E·S·X \\
6  & After months of waiting, a character receives a sudden message that forces them to rethink their entire manuscript. & E·S·R \\
7  & After having recurring dreams, a character gradually accepts that their trust in others has been shattered beyond repair. & I·G·X \\
8  & During a city festival, a sudden rumor quickly spreads and destroys the city’s shared origin story. & E·S·X \\
9  & On a space station where gravity responds to emotions, a character finds anger gradually makes them immobile. & I·G·R \\
10 & Using magic, a character gradually feels each spell weakens their powers, though the belief never fully settles. & I·G·R \\
\addlinespace[1pt]
\multicolumn{3}{l}
{\scriptsize\textit{Abbr.: I/E = internal/external; G/S = gradual/sudden; X/R = irreversible/reversible}}\\
\midrule
\multicolumn{3}{l}{\textbf{\textit{Reorientation}}}\\
11 & After years abroad, a character chooses to return home, seeking the gentle peace they once knew. & V·P·N \\
12 & After writing something late at night, a character calmly walks into the dawn, intent on ending their life. & V·N·N \\
13 & A character accepts a new job offer and starts a routine, feeling neither excitement nor dread. & V·U·N \\
14 & Realizing their childhood longing wasn’t their own, a character lets go of old attachments, hoping for renewal. & V·P·C1 \\
15 & After realizing their emotion affects gravity, a character reconnects with someone from their past to resolve a grudge. & V·P·C9 \\
16 & After a friend’s confession upends everything, a character is irresistibly compelled to seek an unimaginable truth. & IV·P·C4 \\
17 & Yielding to family expectation, a character inherits a shop, sensing their own desires quietly fading. & IV·N·N \\
18 & After receiving a message, a character abandons a lifelong project and starts writing in a genre they resent. & IV·N·C6 \\
19 & After their living situation changes, a character drifts to a new city, adapting to unfamiliar routines without excitement. & IV·U·N \\
20 & After a city festival rumor, a character’s dream of rebuilding fades, and they abandon all further effort. & IV·N·C8 \\
\addlinespace[1pt]
\multicolumn{3}{l}{\scriptsize\textit{Abbr.: V/IV = voluntary/involuntary; P/N/U = positive/negative/neutral; N = not connected; C\# = connected with event constraint \#}}\\
\midrule
\multicolumn{3}{l}{\textbf{\textit{Disruption}}}\\
21 & During a quiet evening at home, an unexpected visitor delivers a shocking news, throwing the household into chaos. & H·S·L \\
22 & After repeated warnings about betrayal, a trusted member is expelled from the group, shattering old bonds. & H·F·L \\
23 & During a national celebration on television, a protester’s sudden action spreads panic throughout the entire country. & H·S·W \\
24 & Ominous weather reports and growing superstition signal disaster before a village becomes gradually isolated from the world. & N·F·L \\
25 & Without warning, an earthquake tears apart neighborhoods and forces families to scatter across a continent. & N·S·W \\
26 & Dead birds and foul smells became more common across the city before authorities declared a state of emergency. & N·F·W \\
27 & A network issue suddenly creates problems for a writer, unexpectedly interrupting the flow of the story. & T·S·L \\
28 & Weeks of ignored security alerts end with a cyberattack that cuts off electricity across the city. & T·F·W \\
29 & At midnight, a secluded old castle fills with unearthly light and its residents instantly vanish. & S·S·L \\
30 & Night after night, strange dreams and omens unsettle the villagers until the entire town disappears. & S·F·W \\
\addlinespace[1pt]
\multicolumn{3}{l}{\scriptsize\textit{Abbr.: H/N/T/S = human/natural/tech/supernatural; F/S = foreshadowed/sudden; L/W = limited/widespread}}\\
\midrule
\multicolumn{3}{l}{\textbf{\textit{Relational Realignment}}}\\
31 & After a heated quarrel, the two brothers refuse to speak, each drifting apart for several months. & SY·D·T \\
32 & When a long-held family secret comes to light, siblings break their silence and stand together from then on. & SY·A·P \\
33 & After failing to reconcile, lifelong friends exchange personal belongings and part ways for a season. & SY·D·T \\
34 & After a failed mediation process, business partners ultimately sever ties permanently and split their shared legacy. & SY·D·P \\
35 & The long-absent member is, if only temporarily, welcomed by the villagers once again, albeit hesitantly. & SY·A·T \\
36 & After a long estrangement, an old friend returns to town, and some quietly welcome them back. & AS·A·T \\
37 & After public humiliation by a mentor, a student destroys a symbol of their apprenticeship and disappears. & AS·D·P \\
38 & After years of silence, a daughter makes a sudden visit, leading to a brief sense of family reunion. & AS·A·T \\
39 & After a scandal, a famous public figure is expelled from the community forever and left utterly isolated. & AS·D·P \\
40 & In the wake of disaster, a newcomer organizes relief, becoming a lasting presence in the entire city. & AS·A·P \\
\addlinespace[1pt]
\multicolumn{3}{l}{\scriptsize\textit{Abbr.: SY/AS = symmetrical/asymmetrical; A/D = alignment/disalignment; T/P = temporary/permanent}}\\
\midrule
\multicolumn{3}{l}{\textbf{\textbf{\textit{Diffusion}}}}\\
41 & After a final conversation at an old meeting place, both parties agree to part and never meet again. & V·S·R \\
42 & Over the years, a close childhood friendship fades as each one finds themselves in distant lands. & IV·G·A \\
43 & One night, during a family gathering, an old feud is suddenly resolved by mutual forgiveness. & V·S·R \\
44 & As memories of a mentor fade, the student finds themselves no longer searching for guidance. & IV·G·A \\
45 & When the final promise is fulfilled at dawn, friends immediately depart, heading into separate unknowns. & V·S·R \\
46 & Over time, the city's once vibrant market empties, and the old merchants quietly move away. & IV·G·A \\
47 & With a single decision at dawn, a character forgives all past wrongs and quietly visits an old friend. & V·S·R \\
48 & After years aboard the generation ship, the crew’s traditions and shared stories gradually fade, leaving only routine survival. & IV·G·A \\
49 & The family abruptly leaves their longtime town behind, closing a chapter in the community’s memory. & V·S·R \\
50 & Over time, a shared dream slips away, and each person lets it go in their own way. & IV·G·A \\
\addlinespace[1pt]
\multicolumn{3}{l}{\scriptsize\textit{Abbr.: V/IV = voluntary/involuntary; S/G = sudden/gradual; R/A = resolution/attrition}}\\
\bottomrule
\end{tabularx}
\end{table}
\vspace{-1em}

\subsection{Style constraints (n=50)}
\vspace{-1.5em}
\begin{table}[H]
\centering
\setlength{\tabcolsep}{1.1pt}
\renewcommand{\arraystretch}{0.97}
\fontsize{7.7pt}{10.1pt}\selectfont
\begin{tabularx}{\textwidth}{@{}p{0.3cm} >{\raggedright\arraybackslash}X >{\raggedleft\arraybackslash}p{1.3cm}@{\hspace{0.15cm}}}
\toprule
\textbf{\#} & \textbf{Constraint} & \textbf{Axes} \\
\midrule
\multicolumn{3}{l}{\textbf{\textit{Write like X}}}\\
1  & Write like Fyodor Dostoevsky. & RL·M·EA \\
2  & Write like Lu Xun. & RL·M·AS \\
3  & Write like Virginia Woolf. & MP·F·EA \\
4  & Write like James Baldwin. & RL·MQ·EA \\
5  & Write like Gabriel García Márquez. & SP·M·GS \\
6  & Write like Octavia Butler. & SP·F·EA \\
7  & Write like Haruki Murakami. & MP·M·AS \\
8  & Write like Jeanette Winterson. & MP·FQ·EA \\
9  & Write like Han Kang. & MP·F·AS \\
10 & Write like Chimamanda Ngozi Adichie. & RL·F·GS \\
\addlinespace[1pt]
\multicolumn{3}{p{\textwidth}}{\scriptsize\textit{abbr.: RL = realist; MP = modernist–postmodernist; SP = speculative; M/F/Q/MQ/FQ = male/female/queer/male+queer/female+queer; EA = euro–american; AS = east asian; GS = global south}}\\
\midrule
\multicolumn{3}{l}{\textbf{\textit{Tone \& Mood}}}\\
11 & Capture a character’s fluid consciousness with vivid sensory detail and nuanced shifts in perception and emotion. & A(VW)·I·V \\
12 & Maintain a cool, melancholic mood, focusing on outward events with abstract and surreal imagery to evoke emotion. & A(HM)·E·A \\
13 & Blend introspective thought and social reality, combining vivid and abstract language for layered, complex scenes. & A(JB)·B·B \\
14 & Describe group interaction and external action, using concrete and balanced expression to sustain a steady mood. & A(CA)·E·B \\
15 & Express psychological tension through internal monologue, using abstract and conceptual language for subtle emotional nuance. & A(HK)·I·A \\
16 & Focus on observable action and outward events, using vivid sensory language and dynamic movement in every scene. & N·E·V \\
17 & Balance inner reflection and outer events, using vivid but ordinary imagery to create a grounded, relatable mood. & N·B·V \\
18 & Objectively describe external situations using abstract, concise language, while minimizing both sensory and dramatic detail. & N·E·A \\
19 & Present both inner feelings and surroundings with abstract, indirect language for a subtle, layered atmosphere. & N·B·A \\
20 & Show a calm, inward-focused mood using vivid, concrete imagery and clear language, avoiding all narrative excess. & N·I·V \\
\addlinespace[1pt]
\multicolumn{3}{p{\textwidth}}{\scriptsize\textit{Abbr.: A(xx) = authorial (VW=Woolf, HM=Murakami, JB=Baldwin, CA=Adichie, HK=Han Kang); N = non-authorial; I/E/B = internal/external/balanced; V/A/B = vivid/abstract/balanced}}\\
\midrule
\multicolumn{3}{l}{\textbf{\textit{Syntax \& Sentence Structure}}}\\
21 & Most sentences are long, structurally complex, and follow standard grammar, prioritizing descriptive narration over dialogue. & C·CV·N \\
22 & Narrative is driven by structurally complex, non-linear sentences, consistently using experimental grammar rather than direct dialogue. & C·E·N \\
23 & Most narration consists of short, direct sentences in standard structure, minimizing dialogue to emphasize exposition. & S·CV·N \\
24 & Short, fragmented sentences break grammatical norms, with narration favored over dialogue in the overall story structure. & S·E·N \\
25 & Dialogue dominates using long, structurally complex sentences and standard grammar, making speech the main storytelling mode. & C·CV·D \\
26 & Dialogue dominates through complex, non-linear sentences with experimental grammar, making speech the primary narrative form. & C·E·D \\
27 & Dialogue drives the narrative, relying on short, direct sentences and standard grammar for a fast, accessible story. & S·CV·D \\
28 & Dialogue dominates through short, fragmented sentences that frequently break grammatical conventions and drive the plot. & S·E·D \\
29 & Dialogue and narration alternate equally, both using standard grammar with mixed complex and simple forms. & B·CV·BA \\
30 & Dialogue and narration appear in nearly equal measure, both frequently using experimental sentence forms and flexible grammar. & B·E·BA \\
\addlinespace[1pt]
\multicolumn{3}{p{\textwidth}}{\scriptsize\textit{Abbr.: C/S/B = complex/simple/balanced; CV/E = conventional/experimental; N/D/BA = narrative/dialogue/balanced}}\\
\midrule
\multicolumn{3}{l}{\textbf{\textit{Temporal Structure}}}\\
31 & Events unfold strictly linearly, compressing years into brief passages, with narration mainly in past tense. & L·C·P \\
32 & The story follows a linear progression, expands single moments over many pages, and uses predominantly the present tense. & L·E·R \\
33 & Linear chronology is used, compressing action to single scenes, with narration almost entirely in the future tense. & L·C·F \\
34 & The story unfolds linearly, expands brief moments into extensive passages, and narration is predominantly in the past tense. & L·E·P \\
35 & Nonlinear structure prevails, compressing long periods with frequent time jumps and narration focused on present-tense events. & N·C·R \\
36 & The narrative is nonlinear, expands single memories into lengthy episodes, and is mainly recounted in the past tense. & N·E·P \\
37 & Nonlinear episodes are compressed into short segments, with narration consistently using the future tense for upcoming events. & N·C·F \\
38 & The nonlinear storyline expands present experiences, drawing out events and emotions with a focus on immediate perception. & N·E·R \\
39 & Fragmented scenes appear out of order, compressing multiple timelines, with narration anchored mainly in the present tense. & FG·C·R \\
40 & Fragmented narrative expands select events in detail, repeatedly anchoring the storytelling in memories and language of the past. & FG·E·P \\
\addlinespace[1pt]
\multicolumn{3}{p{\textwidth}}{\scriptsize\textit{Abbr.: L/N/FG = linear/nonlinear/fragmented; C/E = compressed/expanded; P/R/F = past/present/future}}\\
\midrule
\multicolumn{3}{l}{\textbf{\textit{Narrative Perspective}}}\\
41 & Story is told in first person by a single, reliable narrator, offering subjective depth and emotional intimacy throughout. & 1P·R·S \\
42 & Story is told in first person by a single unreliable narrator, inviting readers’ interpretation of biased events. & 1P·U·S \\
43 & Story is told in second person by a single reliable narrator, immersing readers in events and emotional experience. & 2P·R·S \\
44 & Story is told in third person by a reliable single narrator, providing objective and consistent guidance throughout. & 3P·R·S \\
45 & Story is told in third person by an unreliable single narrator, distorting events and misleading the reader. & 3P·U·S \\
46 & Story is told in first person, alternating multiple reliable narrators to expand subjectivity and narrative scope. & 1P·R·M \\
47 & Story is told in first person by multiple unreliable narrators, each distorting truth and creating fractured, ambiguous reality. & 1P·U·M \\
48 & Story is told in second person by multiple unreliable narrators manipulating truth through shifting roles and conflicting voices. & 2P·U·M \\
49 & Story is told in third person, alternating between multiple reliable narrators, each providing trustworthy and complementary perspectives. & 3P·R·M \\
50 & Story is told in third person by multiple unreliable narrators, presenting distorted versions and erasing truth-lie boundaries. & 3P·U·M \\
\addlinespace[1pt]
\multicolumn{3}{p{\textwidth}}{\scriptsize\textit{Abbr.: 1P/2P/3P = first/second/third person; R/U = reliable/unreliable; S/M = single/multiple}}\\
\bottomrule
\end{tabularx}
\end{table}

\subsection{Character constraints (n=50)}
\vspace{-1em}
\begin{table}[H]
\centering
\setlength{\tabcolsep}{1.2pt}
\renewcommand{\arraystretch}{0.96}
\fontsize{8.5pt}{10pt}\selectfont
\begin{tabularx}{\textwidth}{@{}p{0.3cm} >{\raggedright\arraybackslash}X >{\raggedleft\arraybackslash}p{1.4cm}@{\hspace{0.1cm}}}
\toprule
\textbf{\#} & \textbf{Constraint} & \textbf{Axes} \\
\midrule
\multicolumn{3}{l}{\textbf{\textit{Motive}}}\\
1  & The protagonist is motivated by achievement but torn between high ambition and fear of failure. & D·CO·CF \\
2  & The protagonist is motivated by autonomy, consciously chasing freedom and deliberately forging their own path. & C·CO·F \\
3  & The protagonist is motivated by affiliation, compulsively seeking warmth, belonging, avoiding feeling abandoned or unloved. & D·U·CF \\
4  & The protagonist is motivated by dominance, standing at the center of attention to feel superior. & D·CO·F \\
5  & The protagonist is motivated by nurturance, instinctively devoting their energy to protecting, healing and encouraging. & C·U·F \\
6  & The protagonist is motivated by order, avoiding the chaos with strict routines, acting from habit. & C·U·CF \\
7  & The protagonist is motivated by recognition, fully aware that they thrive on applause and headlines. & D·CO·CF \\
8  & The protagonist is motivated by avoidance, avoiding danger and retreating when facing failure or shame. & D·U·F \\
9  & The protagonist is motivated by counteraction, trying to prove their worth in a healthy direction. & C·CO·CF \\
10 & The protagonist is motivated by understanding, unconsciously striving to understand the world and acquire knowledge. & C·U·F \\
\addlinespace[1pt]
\multicolumn{3}{p{\textwidth}}{\scriptsize\textit{Abbr.: D/C = destructive/constructive; CO/U = conscious/unconscious; CF/F = conflicted/focused}}\\
\midrule
\multicolumn{3}{l}{\textbf{\textit{Social Status}}}\\
11 & The protagonist holds a solid status from birth due to the authority bestowed upon them. & H·I·S \\
12 & The protagonist stands on self-built achievement of high status, yet external changes threaten their status. & H·E·U \\
13 & The protagonist secures middle-class status through effort and skill, maintaining a stable place in society. & M·E·S \\
14 & The protagonist barely maintains the middle-class status they inherited, though it is unstable in society. & M·I·U \\
15 & The protagonist of nobility faces the shadows of the past and the threat of decline. & H·I·U \\
16 & The protagonist of low status lives a stable life, one achieved through their own efforts. & L·E·S \\
17 & The protagonist born into poverty is bound by an unchanging reality, living the same life. & L·I·S \\
18 & The protagonist gains attention through their talent, but their low status makes their life uncertain. & L·E·U \\
19 & The protagonist of the middle class has earned their status, constantly fighting to keep it. & M·E·U \\
20 & The protagonist seeks a future amidst an unstable life and income from a lower-class background. & L·I·U \\
\addlinespace[1pt]
\multicolumn{3}{p{\textwidth}}{\scriptsize\textit{Abbr.: H/M/L = high/middle/low; E/I = earned/inherited; S/U = stable/unstable}}\\
\midrule
\multicolumn{3}{l}{\textbf{\textit{Relational Identity}}}\\
21 & The protagonist engages openly with others, builds trust, and forms bonds based on strong interactions. & C·O \\
22 & The protagonist engages cooperatively and helpfully while defensively controlling the interaction to stay in control. & C·D \\
23 & The protagonist engages quietly, distancing themselves from intimacy and preferring indirect support over deep connections. & C·W \\
24 & The protagonist competes openly, striving to surpass others through noticeable efforts and direct, honest challenges. & M·O \\
25 & The protagonist competes cautiously, torn between the desire to succeed and the fear of failure. & M·D \\
26 & The protagonist competes, distancing themselves from others in pursuit of success but not recognition. & M·W \\
27 & The protagonist competes confidently but manipulates others, leveraging their openness and charm for personal gain. & M·O \\
28 & The protagonist seems sincerely open, but their intentions remain unclear, making them difficult to trust. & A·O \\
29 & The protagonist remains guarded, engaging only when necessary and deflecting others with careful, ambiguous signals. & A·D \\
30 & The protagonist prefers quiet isolation, disconnected from others and uninterested in the world around them. & A·W \\
\addlinespace[1pt]
\multicolumn{3}{p{\textwidth}}{\scriptsize\textit{Abbr.: C/M/A = cooperative/competitive/ambiguous; O/D/W = open/defensive/withdrawn}}\\
\midrule
\multicolumn{3}{l}{\textbf{\textit{Cultural Identity}}}\\
31 & The protagonist fully embraces the dominant culture and is reinforced by institutions, media, and tradition. & MS·M·L \\
32 & The protagonist inherits from ancestors with adopted traditions, expressing multiculturalism within the mainstream society's expectations. & MS·H·L \\
33 & The protagonist lives in between two cultures, never fully accepted or understood by either community. & MG·H·I \\
34 & The protagonist has traditions that are not recognized by society and are disappearing from memory. & MG·M·I \\
35 & The protagonist thrives within a single dominant culture, and their identity is reinforced by institutions. & MS·M·L \\
36 & The protagonist blends global cultures, but their expressions are not read by dominant cultural norms. & MG·H·I \\
37 & The protagonist maintains a single cultural lineage, but is unsupported within the broader social framework. & MG·M·I \\
38 & The protagonist moves between cultures, but society insists on categorizing them as the mainstream group. & MS·H·L \\
39 & The protagonist expresses the dominant culture but hides an invisible identity shaped by their heritage. & MS·H·I \\
40 & The protagonist lives with multiple cultures, one praised in the media, but the other misunderstood. & MG·H·I \\
\addlinespace[1pt]
\multicolumn{3}{p{\textwidth}}{\scriptsize\textit{Abbr.: MS/MG = mainstream/marginalized; M/H = monocultural/hybrid; L/I = legible/illegible}}\\
\midrule
\multicolumn{3}{l}{\textbf{\textit{Embodied Difference}}}\\
41 & The protagonist is an openly nonbinary person whose gender expression is widely accepted in society. & G·AC \\
42 & The protagonist is a disabled person who is often pitied and marginalized despite their ability. & D·SG \\
43 & The protagonist is from a minority ethnic group, their identity erased due to others' indifference. & R·UR \\
44 & The protagonist is an elderly person praised for their wisdom but excluded from decision-making processes. & A·SG \\
45 & The protagonist is a member of the dominant group and is never questioned or “othered.” & U·AC \\
46 & The protagonist is a youthful spirit whose youth is seen as inspiring within their community. & A·AC \\
47 & The protagonist lives invisibly in society despite being gender-nonconforming, ignored in public records and language. & G·UR \\
48 & The protagonist is constantly monitored in society due to racial prejudice, regardless of their actions. & R·SG \\
49 & The protagonist with a cognitive disability is recognized as a valuable contributor and is respected. & D·AC \\
50 & The protagonist blends into the social majority but struggles against the invisibility of being unmarked. & U·UR \\
\addlinespace[1pt]
\multicolumn{3}{p{\textwidth}}{\scriptsize\textit{Abbr.: G/D/R/A/U = gender/disability/race/age/unmarked; AC/SG/UR = accepted/stigmatized/unrecognized}}\\
\bottomrule
\end{tabularx}
\end{table}

\subsection{Setting constraints (n=50)}
\vspace{-1em}
\begin{table}[H]
\centering
\setlength{\tabcolsep}{1.2pt}
\renewcommand{\arraystretch}{0.96}         
\fontsize{8.3 pt}{10.3pt}\selectfont
\begin{tabularx}{\textwidth}{@{}p{0.3cm} >{\raggedright\arraybackslash}X >{\raggedleft\arraybackslash}p{1.6cm}p{0.3cm}}
\toprule
\textbf{\#} & \textbf{Constraint} & \textbf{Axes} \\
\midrule
\multicolumn{3}{l}{\textbf{\textit{Temporal Setting}}}\\
1  & Set in a time when writing, ritual, and early institutions forge enduring cultural foundations. & R·AO \\
2  & Set in a time shaped by sacred knowledge, imperial networks, and slowly shifting frontiers of belief and trade. & R·WFR \\
3  & Set in a time of accelerating change, when new ideas, machines, and ambitions reshape old worlds. & R·WIA \\
4  & Set in a time of total war, collapsing empires, and competing dreams of modernity. & R·SC \\
5  & Set in a present-day or near-future world shaped by digital labor, networked lives, and algorithmic systems. & R·FCN \\
6  & Set in a far future shaped by post-human evolution, unfamiliar ecologies, and fading memories of Earth’s past. & NR·DF \\
7  & Set in a time where causality fractures, and past, present, and future no longer arrive in order. & NR·BS \\
8  & Set in a time shaped by dreams, moods, and symbols, where memory flows deeper than causality. & NR·DT \\
9  & Set in a time so vast that stars rise and die like seconds, and humans flicker like passing thoughts. & NR·CS \\
10 & Set in a time when lives, worlds, or destinies repeat—sometimes exactly, sometimes with a twist. & NR·CR \\
\addlinespace[1pt]
\multicolumn{3}{p{\textwidth}}{\scriptsize\textit{Abbr.: R/NR = realistic/non-realistic; AO = Age of Origins; WFR = Worlds of Faith and Rule; WIA = Worlds in Acceleration; SC = Shattered Century; FCN = Fully Connected Now; DF = Distant Future; BS = Broken Sequence; DT = Dreamtime; CS = Cosmic Scale; CR = Cyclic Return}}\\
\midrule
\multicolumn{3}{l}{\textbf{\textit{Macro Spatial Setting}}}\\
11 & Set in densely constructed spaces where human infrastructure, noise, and social complexity dominate everyday experience. & R·URB \\
12 & Set in cultivated fields, farms, or villages where open landscapes support seasonal rhythms and subsistence life. & R·RUR \\
13 & Set in wooded environments where dense vegetation, biodiversity, and limited visibility shape travel and interaction. & R·FOR \\
14 & Set in high-altitude terrain where isolation, vertical movement, and adaptation to climate define life and architecture. & R·MTN \\
15 & Set in dry, sun-scorched areas with minimal vegetation, scarce water, and extreme diurnal temperature shifts. & R·DES \\
16 & Set in icy, remote zones where cold, wind, and seasonal extremes shape survival and geopolitical activity. & R·POL \\
17 & Set near oceans, lakes, or rivers where water systems define settlement patterns, transportation, and ecological tension. & R·COA \\
18 & Set on alien worlds shaped by unknown atmospheres, strange ecologies, and non-terrestrial natural laws. & NR·XTR \\
19 & Set in digital environments where reality is shaped by code, artificial interaction, and non-physical architecture. & NR·VRT \\
20 & Set in alternate planes of existence ruled by transcendental forces, mythic logic, or timeless ritual. & NR·MYR \\
\addlinespace[1pt]
\multicolumn{3}{p{\textwidth}}{\scriptsize\textit{Abbr.: URB = Urban; RUR = Rural; FOR = Forest; MTN = Mountain; DES = Desert; POL = Polar; COA = Coastal; XTR = Extraterrestrial; VRT = Virtual; MYR = Mythic}}\\
\midrule
\multicolumn{3}{l}{\textbf{\textit{Micro Spatial Setting}}}\\
21 & Set in the interior of a lived-in home, such as a bedroom, kitchen, or shared living area. & R·DOM \\
22 & Set in facilities like schools, factories, or military bases where daily life follows strict organization or control. & R·INS \\
23 & Set in underground or enclosed areas like caves, bunkers, sewers, or hidden chambers, often isolated or secret. & R·SUB \\
24 & Set in spaces designed for movement or passage, such as train stations, highways, ports, or border crossings. & R·TRN \\
25 & Set in ritual or spiritual spaces like temples, altars, shrines, or ancestral enclosures with symbolic significance. & R·SAC \\
26 & Set in places of economic exchange or service, such as markets, shops, offices, or financial institutions. & R·COM \\
27 & Set in hospitals, quarantine zones, labs, or clinics where bodies are treated, monitored, or sequestered. & R·MED \\
28 & Set in digital rooms or artificial environments shaped by code, interaction, and altered perception. & NR·VRI \\
29 & Set in non-logical, symbolic interiors such as looping hallways, floating rooms, or time-shifting apartments. & NR·DLC \\
30 & Set in legendary or magical indoor spaces—cursed castles, sacred vaults, or divine halls shaped by arcane law. & NR·MYS \\
\addlinespace[1pt]
\multicolumn{3}{p{\textwidth}}{\scriptsize\textit{Abbr.: DOM = Domestic; INS = Institutional; SUB = Subterranean; TRN = Transit; SAC = Sacred; COM = Commercial; MED = Medical; VRI = Virtual Interior; DLC = Dreamlike Chamber; MYS = Mythic Structure}}\\
\midrule
\multicolumn{3}{l}{\textbf{\textit{Socio-political Order}}}\\
31 & Set in a world where a powerful central authority enforces strict rules and surveillance to maintain order. & C·S \\
32 & Set in a world where a once-dominant regime is collapsing, creating chaos and shifting power struggles. & C·U \\
33 & Set in a world where religious or ideological laws are absolute, and breaking them is a moral transgression. & C·S \\
34 & Set in a world where machines and systems control society, but human emotions and ethics are fraying. & C·U \\
35 & Set in a world where people live in cooperative harmony, maintaining order through shared values and dialogue. & D·S \\
36 & Set in a world where competing factions each claim authority, but constant disagreements destabilize society. & D·U \\
37 & Set in a world where enduring customs bind society, and decisions emerge through networks of shared practice. & D·S \\
38 & Set in a world where a small community builds its own fragile order on the edge of civilization. & D·U \\
39 & Set in a world where formal institutions have vanished, and survival depends on instinct, alliance, or force. & A·U \\
40 & Set in a world with no governing power, yet operating under alien, ritual, or machinic logics. & A·S \\
\addlinespace[1pt]
\multicolumn{3}{p{\textwidth}}{\scriptsize\textit{Abbr.: C/D/A = Centralized/Distributed/Absent; S/U = Stable/Unstable}}\\
\midrule
\multicolumn{3}{l}{\textbf{\textit{Cultural Context}}}\\
41 & Set in a society where divine will is the ultimate source of law, purpose, and authority. & TH \\
42 & Set in a society where sacred authority is absent, and all norms derive from human reasoning. & AT \\
43 & Set in a society where collective welfare overrides personal choice, and norms are shaped by group will. & C \\
44 & Set in a society where each person is responsible for moral judgment, independent of group consensus. & I \\
45 & Set in a society where moral codes are praised in public but privately ignored by everyone. & HY \\
46 & Set in a society where all know the rules are fake, yet pretend belief sustains stability. & TT \\
47 & Set in a society where norms shift unpredictably, forcing constant adaptation without clear justification. & V \\
48 & Set in a society where rules are applied arbitrarily, and logic never aligns with enforcement. & AR \\
49 & Set in a society where actions follow precedent without question, and justification is neither needed nor allowed. & UQ \\
50 & Set in a society where success defines value, and failure is condemned regardless of intention. & OB \\
\addlinespace[1pt]
\multicolumn{3}{p{\textwidth}}{\scriptsize\textit{Abbr.: TH/AT = Theistic/Atheistic; C/I = Collectivist/Individualist; HY/TT = Hypocritical/Theatrical; V = Volatile; AR = Arbitrary; UQ = Unquestioned; OB = Outcome-based}}\\
\bottomrule
\end{tabularx}
\end{table}

\section{Models and Decoding Parameters}\label{app:models_decoding}
\vspace{-1.4em}
\begin{table}[H]
\centering
\small
\setlength{\tabcolsep}{4pt}
\renewcommand{\arraystretch}{1.1}
\begin{tabularx}{\linewidth}{p{1.3cm} X p{1.4cm} c c c c}
\toprule
\textbf{Abbr.} & \textbf{Full Identifier / Release} & \textbf{Provider} & \textbf{Temp} & \textbf{Top-$p$} & \textbf{reasoning\_effort} & \textbf{verbosity} \\
\midrule
o4mini  & o4-mini-2025-04-16     & OpenAI     & 1.0 & 1.0 & high & \textemdash \\
gpt4.1  & gpt-4.1-2025-04-14     & OpenAI     & 1.0 & 1.0 & \textemdash & \textemdash \\
gpt5    & gpt-5-2025-08-07       & OpenAI     & 1.0 & 1.0 & high & high \\
claude  & claude-opus-4-20250514 & Anthropic  & 1.0 & 1.0 & \textemdash & \textemdash \\
gemini  & gemini-2.5-pro (2025-06-17)         & Google     & 1.0 & 1.0 & \textemdash & \textemdash \\
qwen    & qwen-max-2025-01-25    & Alibaba    & 1.0 & 1.0 & \textemdash & \textemdash \\
\bottomrule
\end{tabularx}
\end{table}
\vspace{-1em}
Models and decoding parameters used in our experiments. For all models, temperature (Temp) and top-$p$ were fixed at 1.0. Vendor-specific controls (reasoning\_effort, verbosity) were set to high when present. For Gemini, an immutable snapshot identifier was not exposed in the API endpoint we used, so we record the model name and the date of access for reproducibility.

\vspace{-0.3em}
\paragraph{Artifact attribution and versioning.}
All models were accessed through the providers’ official APIs. The full identifiers record the provider-released and dated snapshot versions used in our runs. We cite the corresponding official documentation for OpenAI\footnote{https://platform.openai.com/docs/models}, Anthropic\footnote{https://console.anthropic.com/docs/en/home}, Google\footnote{https://ai.google.dev/gemini-api/docs}, and Alibaba Cloud~\footnote{https://www.alibabacloud.com/help/en/model-studio}.

\vspace{-0.3em}
\paragraph{Artifact terms and licensing.}
Our use of proprietary LLMs is governed by the providers’ applicable service terms. OpenAI usage is subject to the OpenAI Services Agreement\footnote{https://openai.com/policies/services-agreement/}. Anthropic usage is subject to Anthropic Commercial Terms of Service\footnote{https://www.anthropic.com/legal/commercial-terms}. Gemini usage is subject to the Gemini API Additional Terms of Service\footnote{https://ai.google.dev/gemini-api/terms}. Qwen usage via Alibaba Cloud is subject to Alibaba Cloud product terms\footnote{https://www.alibabacloud.com/help/en/legal/latest/alibaba-cloud-international-website-product-terms-of-service}.

\section{User Prompts}\label{app:userprompt}
\vspace{-1.4em}

\begin{table}[H]
\centering
\small
\begin{tabularx}{\linewidth}{
  >{\raggedright\arraybackslash}m{0.97\linewidth}
}
\toprule
\multicolumn{1}{c}{\fontsize{10pt}{10pt} \textbf{Experiment 1--1 (single-element, labeled, \textit{K}\,$\ge$\,0)}} \\

\midrule
\begin{minipage}[t]{\linewidth}
\raggedright

\setlength{\baselineskip}{1.1\baselineskip}
As you plan to write a story, identify the specific constraints that would be most useful for writing a single fictional narrative, and explain your reasoning for why each constraint would help write a better narrative.

\medskip
\textbf{Task:}

\noindent - You will be given a list of 50 possible narrative constraints.\\
\noindent - Read through all 50 constraints carefully.\\
\noindent - Select the constraints you consider most useful for writing a fictional narrative.\\
\noindent - You may select as few or as many constraints as you believe are appropriate. There is no minimum or maximum.\\
\noindent - For each selected constraint, explain your reason for choosing it.\\
\noindent - After explaining your individual selections, assess the dynamics among your chosen constraints by explicitly identifying which specific constraints enhance each other and which might interfere with one another. Based on these interactions, evaluate the overall compatibility of your constraint combination and whether it would strengthen or weaken the resulting narrative when applied together in writing.\\
\noindent - There are no restrictions on the length or style of your explanations. Feel free to elaborate as much or as little as you wish.\\
\noindent - You do not need to mention constraints you are not selecting unless you wish to explain why you excluded them.\\
\noindent - List your selections using the specified output format for easy parsing.

\medskip
\textbf{Output Format:}

\noindent - You may select as few or as many constraints as you wish. The order in which you list them does not matter.\\
\noindent - For each, write only the selected constraint as a JSON object, then your reason in the "reason" field.\\
\noindent - Each constraint and its reason must appear as a separate element in a single JSON array containing all elements.\\
\noindent - After listing all selected constraints, include only one paragraph that explains the overall compatibility among all your chosen constraints as a JSON object in the form \{\{"compatibility": "[your explanation]"\}\}, and place it at the end of the array.

\medskip
\textbf{Example Output:} \{example\_lines\_1\_1\}\\
\textbf{Constraint List:} \{constraints\}

\end{minipage}
\\
\bottomrule
\end{tabularx}
\end{table}

\begin{table}[H]
\centering
\small
\begin{tabularx}{\linewidth}{
  >{\raggedright\arraybackslash}m{0.97\linewidth}
}
\toprule
\multicolumn{1}{c}{\fontsize{10pt}{10pt} \textbf{Experiment 1--2 (single-element, labeled, \textit{K}=5)}} \\
\midrule
\begin{minipage}[t]{\linewidth}\raggedright
\setlength{\baselineskip}{1.2\baselineskip}
As you plan to write a story, identify the specific constraints that would be most useful for writing a single fictional narrative, and explain your reasoning for why each constraint would help write a better narrative.

\medskip
\textbf{Task:}

\noindent - You will be given a list of 50 possible narrative constraints.\\
\noindent - Read through all 50 constraints carefully.\\
\noindent - Select exactly 5 constraints you consider most useful for writing a fictional narrative.\\
\noindent - For each selected constraint, explain your reason for choosing it.\\
\noindent - After explaining your individual selections, assess the dynamics among your chosen constraints by explicitly identifying which specific constraints enhance each other and which might interfere with one another. Based on these interactions, evaluate the overall compatibility of your constraint combination and whether it would strengthen or weaken the resulting narrative when applied together in writing.\\
\noindent - There are no restrictions on the length or style of your explanations. Feel free to elaborate as much or as little as you wish.\\
\noindent - You do not need to mention constraints you are not selecting unless you wish to explain why you excluded them.\\
\noindent - List your selections using the specified output format for easy parsing.

\medskip
\textbf{Output Format:}

\noindent - Select exactly 5 constraints. The order in which you list them does not matter.\\
\noindent - For each, write only the selected constraint as a JSON object, then your reason in the "reason" field.\\
\noindent - Each constraint and its reason must appear as a separate element in a single JSON array containing all elements.\\
\noindent - After listing all selected constraints, include only one paragraph that explains the overall compatibility among all your chosen constraints as a JSON object in the form \{\{"compatibility": "[your explanation]"\}\}, and place it at the end of the array.

\medskip
\textbf{Example Output:} \{example\_lines\_1\_2\}\\
\textbf{Constraint List:} \{constraints\}

\end{minipage}
\\
\bottomrule
\end{tabularx}
\end{table}

\begin{table}[H]
\centering
\footnotesize
\begin{tabularx}{\linewidth}{
  >{\raggedright\arraybackslash}m{0.97\linewidth}
}
\toprule
\multicolumn{1}{c}{\fontsize{10pt}{10pt} \textbf{Experiment 2--1 (pooled, unlabeled, \textit{K}\,$\ge$\,0)}} \\
\midrule
\begin{minipage}[t]{\linewidth}\raggedright
\setlength{\baselineskip}{1.2\baselineskip}
As you plan to write a story, identify the specific constraints that would be most useful for writing a single fictional narrative, and explain your reasoning for why each constraint would help write a better narrative.

\medskip
\textbf{Task:}

\noindent - You will be given a list of 200 possible narrative constraints.\\
\noindent - Read through all 200 constraints carefully.\\
\noindent - Select the constraints you consider most useful for writing a fictional narrative.\\
\noindent - You may select as few or as many constraints as you believe are appropriate. There is no minimum or maximum.\\
\noindent - For each selected constraint, explain your reason for choosing it.\\
\noindent - After explaining your individual selections, assess the dynamics among your chosen constraints by explicitly identifying which specific constraints enhance each other and which might interfere with one another. Based on these interactions, evaluate the overall compatibility of your constraint combination and whether it would strengthen or weaken the resulting narrative when applied together in writing.\\
\noindent - There are no restrictions on the length or style of your explanations. Feel free to elaborate as much or as little as you wish.\\
\noindent - You do not need to mention constraints you are not selecting unless you wish to explain why you excluded them.\\
\noindent - List your selections using the specified output format for easy parsing.

\medskip
\textbf{Output Format:}

\noindent - You may select as few or as many constraints as you wish. The order in which you list them does not matter.\\
\noindent - For each, write only the selected constraint as a JSON object, then your reason in the "reason" field.\\
\noindent - Each constraint and its reason must appear as a separate element in a single JSON array containing all elements.\\
\noindent - After listing all selected constraints, include only one paragraph that explains the overall compatibility among all your chosen constraints as a JSON object in the form \{\{"compatibility": "[your explanation]"\}\}, and place it at the end of the array.

\medskip
\textbf{Example Output:} \{example\_lines\_2\_1\}\\
\textbf{Constraint List:} \{constraints\}

\end{minipage}
\\
\bottomrule
\end{tabularx}\label{app:userprompt12}
\end{table}
\vspace{-3em}

\begin{table}[H]
\centering
\footnotesize
\begin{tabularx}{\linewidth}{
  >{\raggedright\arraybackslash}m{0.97\linewidth}
}
\toprule
\multicolumn{1}{c}{\fontsize{10pt}{10pt} \textbf{Experiment 2--2 (pooled, unlabeled, \textit{K}=20)}} \\
\midrule
\begin{minipage}[t]{\linewidth}\raggedright
\setlength{\baselineskip}{1.2\baselineskip}
As you plan to write a story, identify the specific constraints that would be most useful for writing a single fictional narrative, and explain your reasoning for why each constraint would help write a better narrative.

\medskip
\textbf{Task:}

\noindent - You will be given a list of 200 possible narrative constraints.\\
\noindent - Read through all 200 constraints carefully.\\
\noindent - Select exactly 20 constraints you consider most useful for writing a fictional narrative.\\
\noindent - For each selected constraint, explain your reason for choosing it.\\
\noindent - After explaining your individual selections, assess the dynamics among your chosen constraints by explicitly identifying which specific constraints enhance each other and which might interfere with one another. Based on these interactions, evaluate the overall compatibility of your constraint combination and whether it would strengthen or weaken the resulting narrative when applied together in writing.\\
\noindent - There are no restrictions on the length or style of your explanations. Feel free to elaborate as much or as little as you wish.\\
\noindent - You do not need to mention constraints you are not selecting unless you wish to explain why you excluded them.\\
\noindent - List your selections using the specified output format for easy parsing.

\medskip
\textbf{Output Format:}

\noindent - Select exactly 20 constraints. The order in which you list them does not matter.\\
\noindent - For each, write only the selected constraint as a JSON object, then your reason in the "reason" field.\\
\noindent - Each constraint and its reason must appear as a separate element in a single JSON array containing all elements.\\
\noindent - After listing all selected constraints, include only one paragraph that explains the overall compatibility among all your chosen constraints as a JSON object in the form \{\{"compatibility": "[your explanation]"\}\}, and place it at the end of the array.

\medskip
\textbf{Example Output:} \{example\_2\_2\}\\
\textbf{Constraint List:} \{constraints\}

\end{minipage}
\\
\bottomrule
\end{tabularx}
\end{table}

\begin{table}[H]
\centering
\footnotesize
\begin{tabularx}{\linewidth}{
  >{\raggedright\arraybackslash}m{0.97\linewidth}
}
\toprule
\multicolumn{1}{c}{\fontsize{10pt}{10pt} \textbf{Experiment 3 (element-blocked, labeled with quotas; \textit{K}=20, \textit{k}=5 per element)}} \\
\midrule
\begin{minipage}[t]{\linewidth}\raggedright
\setlength{\baselineskip}{1.2\baselineskip}
As you plan to write a story, identify the specific constraints that would be most useful for writing a single fictional narrative, and explain your reasoning for why each constraint would help write a better narrative.

\medskip
\textbf{Task:}

\noindent - You will complete this task for four core narrative elements: event, style, character, and setting.\\
\noindent - For each element, you will be given a list of 50 possible narrative constraints.\\
\noindent - Read through all 50 constraints for each element carefully.\\
\noindent - For each element, select exactly 5 constraints that you believe are most useful for writing a fictional narrative.\\
\noindent - For each selected constraint, explain your reason for choosing it.\\
\noindent - After explaining your individual selections, assess the dynamics among your chosen constraints by explicitly identifying which specific constraints enhance each other and which might interfere with one another. Based on these interactions, evaluate the overall compatibility of your constraint combination and whether it would strengthen or weaken the resulting narrative when applied together in writing.\\
\noindent - There are no restrictions on the length or style of your explanations. Feel free to elaborate as much or as little as you wish.\\
\noindent - List your selections for each element using the specified output format for easy parsing.

\medskip
\textbf{Output Format:}

\noindent - For each element, list exactly 5 constraints. The order in which you list them does not matter.\\
\noindent - For each, write only the selected constraint as a JSON object, then your reason in the "reason" field.\\
\noindent - Each constraint and its reason must appear as a separate element in a single JSON array containing all elements.\\
\noindent - After listing all selected constraints, include only one paragraph that explains the overall compatibility among all your chosen constraints as a JSON object in the form \{\{"compatibility": "[your explanation]"\}\}, and place it at the end of the array.

\medskip
\textbf{Example Output:} \{example\_lines\_3\}\\
\textbf{Constraint List:} \{constraints\}

\end{minipage}
\\
\bottomrule
\end{tabularx}
\end{table}

\section{Instruction Type}\label{app:system}
\vspace{-1em}
\begin{table}[H]
\renewcommand{\arraystretch}{1.3}
\centering
\begin{tabularx}{\textwidth}{
  >{\centering\arraybackslash}m{2.9cm}
  >{\raggedright\arraybackslash}m{12.6cm}
}
\toprule
\textbf{Instruction Type} & \textbf{Description} \\
\midrule
Basic & You are a writer. Your task is to write narratives when requested. Your goal is to write complete narratives that fulfill the given requirements. \\
\midrule
Quality-focused & You are a highly skilled writer known for technical excellence and flawless execution of storytelling fundamentals. You write stories with precise character development, well-structured plots, polished prose, and carefully integrated themes. Your goal is to write stories of the highest quality through careful refinement and technical mastery. \\
\midrule
Creativity-focused & You are an innovative writer celebrated for creating completely original and unexpected narratives. You excel at breaking conventional storytelling rules and exploring new creative possibilities. Your strength lies in developing unique characters, unusual plot structures, or experimental styles that surprise readers. Your goal is to create narratives that are unlike anything that has been written before, pushing the boundaries of what stories can be through creative experimentation. \\
\bottomrule
\end{tabularx}
\label{tab:system-prompts}
\end{table}
\vspace{1em}


\section{Statistical Modeling, Inference, and Diagnostics}\label{app:stats}

This appendix reports the full outcome definitions, model specifications, inference procedures, and diagnostics that are referenced in the main Results section.

\subsection{Outcome Definitions and Condition-Contrast Modeling}\label{app:stats_outcome}

\paragraph{Outcome \& modeling.}
For each unit–category $(u,c)$ we compute the within-unit selection share
$s_{uc}=y_{uc}/K_u$ and control for supply via the supply share
$p_{uc}=n_{uc}/N_u$ (covariate adjustment).
Category-wise percentage-point differences in selection shares (in pp) between conditions are estimated using both OLS and $K$-weighted WLS (weights $=K_u$) with run-clustered SEs; ~\autoref{tab:condition_contrasts} reports the $K$-weighted WLS estimates.
Heterogeneity is assessed via Wald tests on $D\times$model and $D\times$instruction type
interactions, where $D$ encodes the planned contrasts
(1--2 vs.\ 1--1, 2--2 vs.\ 2--1, 3 vs.\ 1--2, 3 vs.\ 2--2).
We report two-sided $p$-values with 95\% CIs and adjust for multiple testing across the family of pairwise contrasts using BH--FDR.

\paragraph{Power analysis for Stage~2 replication count.}
To choose the Stage~2 replication count for Experiment~2--2, we conducted an RR-based power analysis targeting stable detection of moderate composition shifts. Using an a priori 80th-percentile coverage criterion across model$\times$instruction-type strata and accounting for run-level exposure and overdispersion (median \(K\approx 20\); \(\phi_{p90}=1.00\)), the required runs per group were 95 for RR\(=1.20\) at the element level and 154 for RR\(=1.50\) at the category level. We therefore set \(R=160\).

\subsection{Poisson GEE Specifications for Composition Models}\label{app:stats_gee}

\subsubsection{Element-Level Composition (Experiment 2--2)}\label{app:stats_gee_element}

\paragraph{Model \& contrasts.}
As specified in Methodology section, we analyze run–element counts with a run–clustered Poisson GEE using element effects and element$\times$(model, instruction type) interactions (no intercept). We do not include model$\times$instruction type (or higher-order) interactions, so prompt contrasts are averaged over models and vice versa. We report (i) element rate ratios (RRs) relative to \emph{Event}, which we use as an arbitrary reference category, and (ii) within–element pairwise RRs for model and for instruction type.

\subsubsection{Category-Level Composition (Experiment 2--2)}\label{app:stats_gee_category}

\paragraph{Model \& contrasts.}
For each element, we analyze run--category counts with a run--clustered Poisson GEE using category effects and category$\times$(model, instruction type) interactions (no intercept). We do not include model$\times$instruction type (or higher--order) interactions; consequently, instruction type contrasts are averaged over models (and model contrasts over prompts). We report (i) category rate ratios (RRs) relative to a baseline category within each element and (ii) within--category pairwise RRs for model and for instruction type.

\subsection{Inference, Standard Errors, and Reporting Conventions}\label{app:stats_inference}

\subsubsection{Element-Level Inference and Reporting}\label{app:stats_inference_element}

\paragraph{Inference \& reporting.}
Our specification uses the exposure offset $\log K$. Inference uses Wald $\chi^2(1)$ tests on log--rate contrasts with robust covariance, and 95\% CIs are $\exp(\hat\theta \pm 1.96\,\mathrm{SE})$.
Tables indicate the number of run clusters.
We report pairwise differences only when BH--FDR $q<.05$ and $|\Delta\%|\ge 10$.

\subsubsection{Category-Level Inference and Reporting}\label{app:stats_inference_category}

\paragraph{Inference \& reporting.}
Our specification uses the exposure offset $\log K_{\mathrm{elem}}$. Inference uses Wald $\chi^2(1)$ tests on log--rate contrasts with robust covariance, and 95\% CIs are $\exp(\hat\theta \pm 1.96\,\mathrm{SE})$.
Tables indicate the number of run clusters.
We report pairwise differences only when BH--FDR $q<.05$ and $|\Delta\%|\ge 10$.
In the main text, ~\autoref{tab:category_main} summarizes large category shifts vs.\ within-element baselines and reports only categories with $p<.05$ and $|\Delta|\ge 50\%$.

\subsection{Model Adequacy and Robustness Checks}\label{app:stats_adequacy}

\paragraph{Model adequacy.}
Across all Poisson GEE fits, dispersion diagnostics were below unity (elements: Pearson $\chi^2/\mathrm{df}=0.567$, deviance/df $=0.611$; categories: $\chi^2/\mathrm{df}=0.402$–$0.664$, deviance/df $=0.454$–$0.740$), with many run clusters (elements: $n=2{,}880$; categories: $n=2{,}793$–$2{,}880$; runs with $K_{\mathrm{elem}}=0$ excluded from the corresponding element-specific \emph{category} models); we therefore report run-clustered robust (sandwich) SEs and fit Poisson GEEs with an exchangeable working correlation by default; when the GEE fit failed or produced invalid covariance estimates, we refit the model using an independence working correlation; if that specification also failed, we used a Poisson GLM with run-clustered robust covariance as a fallback.

\subsection{Axis-Level Permutation Test and Axis Aggregation}\label{app:stats_axis}

\paragraph{Model \& test.}
Within Experiment~2--2, we assess constraint\hyp{}level over\hyp{} or under\hyp{}selection via a Monte Carlo permutation test stratified by \(\text{model}\times\text{instruction type}\times\text{element}\times\text{category}\). The null assumes exchangeability across constraints conditional on each run’s selection budget \(K_u\) and pool composition. For each constraint \(c\), we compute the observed total \(Y_c=\sum_u y_{uc}\) and the supply\hyp{}adjusted expectation \(\mathbb{E}[Y_c]=\sum_u K_u\,(n_{c,u}/N_u)\). We report \(\mathrm{share}_{\mathrm{obs}}=Y_c/\sum_u K_u\), \(\mathrm{share}_{\mathrm{exp}}=\mathbb{E}[Y_c]/\sum_u K_u\), \(\mathrm{RD}_{\mathrm{share}}=\mathrm{share}_{\mathrm{obs}}-\mathrm{share}_{\mathrm{exp}}\), and a smoothed observed-to-expected ratio \(\mathrm{Obs/Exp}=(Y_c+0.5)/(\mathbb{E}[Y_c]+0.5)\); direction is defined by \(\mathrm{share}_{\mathrm{obs}}\) vs.\ \(\mathrm{share}_{\mathrm{exp}}\).

\paragraph{Inference \& reporting.}
Two\hyp{}sided \(p\)\hyp{}values (and one\hyp{}sided \(p_{\text{over}}, p_{\text{under}}\)) come from \(B{=}2000\) permutations with a +1 correction; multiplicity is controlled within stratum by BH--FDR on \(p_{\text{two}}\) (significance at \(q\le .10\); fallback \(p_{\text{two}}\le .05\) in degenerate strata). Axis\hyp{}level summaries map constraint signals to axis annotations and aggregate them on a shared observed/expected scale. For the \emph{global axis baseline}, we pool \((Y_c,\mathbb{E}[Y_c])\) across all strata and then aggregate to axes, reporting observed and expected axis shares and their ratio. For \emph{instruction-type contrasts}, we restrict to significant constraints within each stratum, propagate their over/under direction to axes, and compute within-direction axis shares along with enrichment ratios defined as \(\mathrm{Share}/\mathrm{Global}\), where \(\mathrm{Global}\) denotes the pooled baseline share within the same direction (pooled across instruction types and models). Summary tables apply minimum-support thresholds (MIN\_SUPPORT) and Top-\(K\)/union rules for visualization. We set MIN\_SUPPORT = 3, reporting an axis-direction only when it is supported by at least three distinct significant constraints, to reduce single-constraint artifacts and stabilize the descriptive summaries.

\section{Category-Level Selection Rate Ratios vs.\ Within-Element Baseline Category}\label{app:cat_main_rr}
\vspace{-2.5em}

\begin{table}[H]
\setlength{\tabcolsep}{5pt}
\renewcommand{\arraystretch}{1.2}
\begin{tabularx}{\textwidth}{@{} c >{\raggedright\arraybackslash}X c c c @{}}
\toprule
\textbf{Element} & \textbf{Category} & \textbf{RR [95\% CI]} & \textbf{$p$} & \textbf{N runs} \\
\midrule
\multirow{5}{*}{\textit{Event}} & \textit{Diffusion} (baseline) & 1.00 [1.00, 1.00] & - & 2880 \\
 & \textit{Disruption} & 1.23 [1.16, 1.31] & < .001 & 2880 \\
 & \textit{Epistemological Transformation} & 2.15 [2.03, 2.27] & < .001 & 2880 \\
 & \textit{Relational Realignment} & 0.62 [0.58, 0.68] & < .001 & 2880 \\
 & \textit{Reorientation} & 1.50 [1.41, 1.59] & < .001 & 2880 \\
\midrule
\multirow{5}{*}{\textit{Style}} & \textit{Narrative perspective} (baseline) & 1.00 [1.00, 1.00] & - & 2880 \\
 & \textit{Syntax \& Sentence Structure} & 0.61 [0.59, 0.64] & < .001 & 2880 \\
 & \textit{Temporal Structure} & 0.97 [0.95, 1.00] & .029 & 2880 \\
 & \textit{Tone \& Mood} & 2.27 [2.23, 2.32] & < .001 & 2880 \\
 & \textit{Write like X} & 0.35 [0.33, 0.38] & < .001 & 2880 \\
\midrule
\multirow{5}{*}{\textit{Character}} & {\textit{Cultural Identity} (baseline)} & 1.00 [1.00, 1.00] & - & 2880 \\
 & \textit{Embodied Difference} & 0.58 [0.54, 0.62] & < .001 & 2880 \\
 & \textit{Motive} & 2.09 [2.00, 2.18] & < .001 & 2880 \\
 & \textit{Relational Identity} & 1.05 [1.00, 1.11] & .059 & 2880 \\
 & \textit{Social Status} & 0.59 [0.55, 0.64] & < .001 & 2880 \\
\midrule
\multirow{5}{*}{\textit{Setting}} & \textit{Cultural context} (baseline) & 1.00 [1.00, 1.00] & - & 2880 \\
 & \textit{Macro spatial setting} & 1.18 [1.12, 1.24] & < .001 & 2880 \\
 & \textit{Micro spatial setting} & 1.18 [1.12, 1.24] & < .001 & 2880 \\
 & \textit{Socio-political order} & 1.10 [1.04, 1.16] & .001 & 2880 \\
 & \textit{Temporal setting} & 1.57 [1.49, 1.64] & < .001 & 2880 \\
\bottomrule
\end{tabularx}
\end{table}
\vspace{-2em}
Category-level selection rate ratios vs.\ the within-element baseline category (pooled across models/prompts), with exposure offset $\log K_{\mathrm{elem}}$. Runs with $K_{\mathrm{elem}}=0$ are excluded from the corresponding element-specific category models; run clusters therefore vary by element (2,793--2,880).
\vspace{23em}

\normalsize

\section{Top-15 Globally Over- or Under-selected Axes}\label{app:global_axis_enrichment}
\vspace{-1em}
\begin{table}[H]
\centering
\normalsize
\setlength{\tabcolsep}{4pt}
\fontsize{10.5pt}{11.5pt}
\renewcommand{\arraystretch}{1.1}
\begin{tabularx}{\linewidth}{l X r r r r}
\toprule
\textbf{Category} & \textbf{Axis} & \textbf{Obs} & \textbf{Obs/Exp ($\times$)} & \textbf{Obs (\%)} & \textbf{Exp (\%)} \\
\midrule
\multicolumn{6}{l}{\textbf{Global --- Over-selected}}\\
\textit{Reorientation} & \textit{Connected (E4)} & 324 & 3.89 & 0.56 & 0.14 \\
\textit{Macro spatial setting} & \textit{Urban Built Environments} & 508 & 3.66 & 0.88 & 0.24 \\
\textit{Temporal setting} & \textit{The Fully Connected Now} & 523 & 2.74 & 0.91 & 0.33 \\
\textit{Micro spatial setting} & \textit{Dreamlike or Surreal Chambers} & 357 & 2.52 & 0.62 & 0.25 \\
\textit{Cultural context} & \textit{Theatrical Order} & 611 & 2.36 & 1.06 & 0.45 \\
\textit{Reorientation} & \textit{Connected (E1)} & 189 & 2.27 & 0.33 & 0.14 \\
\textit{Syntax \& Sentence Structure} & \textit{Balanced} & 488 & 2.07 & 0.85 & 0.41 \\
\textit{Reorientation} & \textit{Positive} & 684 & 2.06 & 1.19 & 0.58 \\
\textit{Tone \& Mood} & \textit{Authorial (Virginia Woolf)} & 578 & 1.99 & 1.00 & 0.50 \\
\textit{Tone \& Mood} & \textit{Authorial (James Baldwin)} & 834 & 1.92 & 1.45 & 0.75 \\
\textit{Temporal setting} & \textit{The Broken Sequence} & 352 & 1.85 & 0.61 & 0.33 \\
\textit{Micro spatial setting} & \textit{Domestic Interior Spaces} & 250 & 1.76 & 0.43 & 0.25 \\
\textit{Embodied Difference} & \textit{Race-Marked} & 206 & 1.69 & 0.36 & 0.21 \\
\textit{Temporal setting} & \textit{Worlds in Acceleration} & 322 & 1.69 & 0.56 & 0.33 \\
\textit{Disruption} & \textit{Technological} & 235 & 1.59 & 0.41 & 0.26 \\
\addlinespace[2pt]
\multicolumn{6}{l}{\textbf{Global --- Under-selected}}\\
\textit{Temporal setting} & \textit{Age of Origins} & 32 & 0.17 & 0.06 & 0.33 \\
\textit{Cultural context} & \textit{Theistic} & 45 & 0.18 & 0.08 & 0.45 \\
\textit{Temporal Structure} & \textit{Future} & 47 & 0.19 & 0.08 & 0.43 \\
\textit{Reorientation} & \textit{Connected (E8)} & 17 & 0.21 & 0.03 & 0.14 \\
\textit{Temporal setting} & \textit{Worlds of Faith and Rule} & 41 & 0.22 & 0.07 & 0.33 \\
\textit{Reorientation} & \textit{Neutral} & 40 & 0.24 & 0.07 & 0.29 \\
\textit{Macro spatial setting} & \textit{Mountainous Regions} & 34 & 0.25 & 0.06 & 0.24 \\
\textit{Cultural Identity} & \textit{Legible} & 69 & 0.25 & 0.12 & 0.48 \\
\textit{Cultural context} & \textit{Unquestioned Precedent} & 67 & 0.26 & 0.12 & 0.45 \\
\textit{Macro spatial setting} & \textit{Polar and Glacial Frontiers} & 36 & 0.26 & 0.06 & 0.24 \\
\textit{Macro spatial setting} & \textit{Deserts and Arid Zones} & 40 & 0.29 & 0.07 & 0.24 \\
\textit{Temporal setting} & \textit{The Shattered Century} & 60 & 0.31 & 0.10 & 0.33 \\
\textit{Reorientation} & \textit{Negative} & 105 & 0.32 & 0.18 & 0.58 \\
\textit{Relational Identity} & \textit{Competitive} & 146 & 0.33 & 0.25 & 0.77 \\
\textit{Micro spatial setting} & \textit{Mythic or Enchanted Structures} & 47 & 0.33 & 0.08 & 0.25 \\
\bottomrule
\end{tabularx}
\end{table}
\vspace{-1em}
Top-15 globally \emph{over}- or \emph{under}-selected axes across all runs (descriptive; no significance filtering). Obs and Exp are aggregated observed and supply-adjusted expected counts; Obs/Exp $=(\mathrm{Obs}+0.5)/(\mathrm{Exp}+0.5)$. Obs (\%) and Exp (\%) report observed and expected axis shares. Values rounded.
\vspace{11em}

\section{Top-5 Over- or Under-selected Axes by Instruction Type}\label{app:axis_top5_instruction_type}
\vspace{-1em}
\begin{table}[H]
\centering
\setlength{\tabcolsep}{4pt}
\small
\renewcommand{\arraystretch}{1.3}
\begin{tabularx}{\linewidth}{l X r r r r}
\toprule
\textbf{Category} & \textbf{Axis} & \textbf{Support} & \textbf{Enrich ($\times$)} & \textbf{Share (\%)} & \textbf{Global (\%)} \\
\midrule
\multicolumn{6}{l}{\textbf{Basic — Over-selected}}\\
\textit{Embodied Difference} & \textit{Disability-Marked} & 3 & 3.23 & 2.52 & 0.78 \\
\textit{Micro spatial setting} & \textit{Domestic Interior Spaces} & 4 & 3.20 & 2.50 & 0.78 \\
\textit{Tone \& Mood} & \textit{Authorial (James Baldwin)} & 5 & 3.15 & 2.46 & 0.78 \\
\textit{Tone \& Mood} & \textit{Authorial (Virginia Woolf)} & 5 & 3.15 & 2.46 & 0.78 \\
\textit{Temporal setting} & \textit{Realistic} & 12 & 3.12 & 4.88 & 1.56 \\
\addlinespace[2pt]
\multicolumn{6}{l}{\textbf{Basic — Under-selected}}\\
\textit{Write like X} & \textit{East Asian} & 3 & 3.42 & 3.80 & 1.11 \\
\textit{Write like X} & \textit{Male} & 4 & 3.04 & 5.06 & 1.67 \\
\textit{Tone \& Mood} & \textit{Authorial (Chimamanda Adichie)} & 4 & 2.66 & 1.48 & 0.56 \\
\textit{Temporal setting} & \textit{The Dreamtime} & 5 & 2.28 & 1.27 & 0.56 \\
\textit{Tone \& Mood} & \textit{Authorial (Haruki Murakami)} & 5 & 2.28 & 1.27 & 0.56 \\
\midrule
\multicolumn{6}{l}{\textbf{Quality — Over-selected}}\\
\textit{Reorientation} & \textit{Not Connected} & 3 & 3.69 & 2.88 & 0.78 \\
\textit{Tone \& Mood} & \textit{Authorial (James Baldwin)} & 5 & 3.37 & 2.63 & 0.78 \\
\textit{Tone \& Mood} & \textit{Authorial (Virginia Woolf)} & 5 & 3.37 & 2.63 & 0.78 \\
\textit{Macro spatial setting} & \textit{Aquatic and Coastal Environments} & 4 & 3.32 & 2.60 & 0.78 \\
\textit{Micro spatial setting} & \textit{Domestic Interior Spaces} & 4 & 3.28 & 2.56 & 0.78 \\
\addlinespace[2pt]
\multicolumn{6}{l}{\textbf{Quality — Under-selected}}\\
\textit{Write like X} & \textit{East Asian} & 4 & 4.50 & 5.00 & 1.11 \\
\textit{Write like X} & \textit{Modernist-Postmodernist} & 3 & 3.38 & 3.75 & 1.11 \\
\textit{Write like X} & \textit{Male} & 4 & 3.00 & 5.00 & 1.67 \\
\textit{Tone \& Mood}\ & \textit{Authorial (Chimamanda Adichie)} & 4 & 2.51 & 1.39 & 0.56 \\
\textit{Write like X} & \textit{Euro-American} & 3 & 2.25 & 3.75 & 1.67 \\
\midrule
\multicolumn{6}{l}{\textbf{Creativity — Over-selected}}\\
\textit{Macro spatial setting} & \textit{Extraterrestrial Terrain} & 4 & 3.22 & 2.52 & 0.78 \\
\textit{Macro spatial setting} & \textit{Otherworldly or Mythic Realms} & 4 & 3.22 & 2.52 & 0.78 \\
\textit{Cultural context} & \textit{Volatile Norms} & 5 & 3.11 & 2.43 & 0.78 \\
\textit{Reorientation} & \textit{Connected} (E9) & 5 & 3.11 & 2.43 & 0.78 \\
\textit{Temporal setting} & \textit{The Dreamtime} & 5 & 3.11 & 2.43 & 0.78 \\
\addlinespace[2pt]
\multicolumn{6}{l}{\textbf{Creativity — Under-selected}}\\
\textit{Write like X} & \textit{Global South} & 3 & 3.10 & 1.72 & 0.56 \\
\textit{Write like X} & \textit{Realist} & 11 & 2.84 & 6.32 & 2.22 \\
\textit{Write like X} & \textit{Queer} & 3 & 2.84 & 3.16 & 1.11 \\
\textit{Write like X} & \textit{Male} & 8 & 2.76 & 4.60 & 1.67 \\
\textit{Cultural context} & \textit{Collectivist} & 5 & 2.41 & 1.34 & 0.56 \\
\bottomrule
\end{tabularx}
\end{table}
\vspace{-1em}
Top-5 Over- or Under-selected axes by instruction type. Support = pooled count of significant constraints mapped to the axis (summed across models). Enrich = Share / Global, where Share = axis share within direction and Global = pooled baseline share within the same direction (computed over significant constraints, pooled across instruction types and models). Values rounded.

\newpage
\section{Model-Stratified Axis Summaries}\label{app:axis_model_summaries}

\subsection{Axes Shared Across Models}\label{app:axis_model_common}
\vspace{-1.2em}

\begin{table}[H]
\centering
\footnotesize
\setlength{\tabcolsep}{4pt}
\renewcommand{\arraystretch}{1.12}
\begin{tabularx}{\linewidth}{p{2.6cm} X c c c c c c}
\toprule
\textbf{Category} & \textbf{Axis} &
\textbf{claude} & \textbf{gemini} & \textbf{gpt4.1} & \textbf{gpt5} & \textbf{o4mini} & \textbf{qwen} \\
\midrule
Motive & Conscious
& \hi & \hi & \hi\ \lo & \hi & \hi & \hi\ \lo \\
Motive & Constructive
& \hi & \hi & \hi & \hi & \hi\ \lo & \hi\ \lo \\
Temporal setting & Realistic
& \hi & \hi & \hi\ \lo & \hi & \hi & \hi \\
\bottomrule
\end{tabularx}
\end{table}

\vspace{-1em}
Axes common to all models (Top-20 per model via over/under union; axes with a uniform direction across models excluded). Cells mark per-model direction: \hi{}=over-selected, \lo{}=under-selected. A paired mark (\hi{}\ \lo{}) indicates that the direction flips across instruction types (instruction-contingent).
\vspace{1em}

\subsection{Top-5 Over- or Under-selected Axes by Model (claude, gemini, gpt4.1)}\label{app:axis_top5_by_model_claude_gemini_gpt4.1}
\vspace{-1em}
\begin{table}[H]
\centering
\small
\setlength{\tabcolsep}{4pt}
\renewcommand{\arraystretch}{1.2}
\label{app:axis_top5_by_model_a}
\begin{tabularx}{\linewidth}{l X r r r r}
\toprule
\textbf{Category} & \textbf{Axis} & \textbf{Support} & \textbf{Enrich ($\times$)} & \textbf{Share (\%)} & \textbf{Global (\%)} \\
\midrule
\multicolumn{6}{l}{\textbf{claude — Over-selected}}\\
\textit{Temporal setting} & \textit{Realistic} & 4 & 3.61 & 5.63 & 1.56 \\
\textit{Reorientation} & \textit{Connected (E4)} & 3 & 3.37 & 2.63 & 0.78 \\
\textit{Tone \& Mood} & \textit{Authorial (James Baldwin)} & 3 & 3.37 & 2.63 & 0.78 \\
\textit{Tone \& Mood} & \textit{Authorial (Virginia Woolf)} & 3 & 3.37 & 2.63 & 0.78 \\
\textit{Narrative perspective} & \textit{Unreliable} & 4 & 2.98 & 9.30 & 3.12 \\
\addlinespace[2pt]
\multicolumn{6}{l}{\textbf{claude — Under-selected}}\\
\textit{Reorientation} & \textit{Neutral} & 6 & 2.02 & 2.25 & 1.11 \\
\textit{Macro spatial setting} & \textit{Deserts and Arid Zones} & 3 & 2.02 & 1.12 & 0.56 \\
\textit{Macro spatial setting} & \textit{Mountainous Regions} & 3 & 2.02 & 1.12 & 0.56 \\
\textit{Macro spatial setting} & \textit{Polar and Glacial Frontiers} & 3 & 2.02 & 1.12 & 0.56 \\
\textit{Micro spatial setting} & \textit{Sacred Grounds} & 3 & 2.02 & 1.12 & 0.56 \\
\midrule
\multicolumn{6}{l}{\textbf{gemini — Over-selected}}\\
\textit{Embodied Difference} & \textit{Stigmatized} & 4 & 2.84 & 4.44 & 1.56 \\
\textit{Temporal setting} & \textit{Realistic} & 4 & 2.84 & 4.44 & 1.56 \\
\textit{Cultural context} & \textit{Theatrical Order} & 3 & 2.80 & 2.19 & 0.78 \\
\textit{Micro spatial setting} & \textit{Dreamlike or Surreal Chambers} & 3 & 2.80 & 2.19 & 0.78 \\
\textit{Relational Identity} & \textit{Defensive} & 3 & 2.80 & 2.19 & 0.78 \\
\addlinespace[2pt]
\multicolumn{6}{l}{\textbf{gemini — Under-selected}}\\
\textit{Cultural Identity} & \textit{Legible} & 12 & 1.77 & 3.93 & 2.22 \\
\textit{Relational Identity} & \textit{Competitive} & 12 & 1.77 & 3.93 & 2.22 \\
\textit{Reorientation} & \textit{Neutral} & 6 & 1.77 & 1.97 & 1.11 \\
\textit{Cultural context} & \textit{Atheistic} & 3 & 1.77 & 0.98 & 0.56 \\
\textit{Cultural context} & \textit{Collectivist} & 3 & 1.77 & 0.98 & 0.56 \\
\midrule
\multicolumn{6}{l}{\textbf{gpt4.1 — Over-selected}}\\
\textit{Temporal setting} & \textit{Realistic} & 4 & 3.82 & 5.97 & 1.56 \\
\textit{Reorientation} & \textit{Connected (E1)} & 3 & 3.66 & 2.86 & 0.78 \\
\textit{Reorientation} & \textit{Connected (E4)} & 3 & 3.66 & 2.86 & 0.78 \\
\textit{Tone \& Mood} & \textit{Authorial (James Baldwin)} & 3 & 3.66 & 2.86 & 0.78 \\
\textit{Tone \& Mood} & \textit{Authorial (Virginia Woolf)} & 3 & 3.66 & 2.86 & 0.78 \\
\addlinespace[2pt]
\multicolumn{6}{l}{\textbf{gpt4.1 — Under-selected}}\\
\textit{Temporal Structure} & \textit{Future} & 6 & 3.94 & 4.38 & 1.11 \\
\textit{Reorientation} & \textit{Connected (E6)} & 3 & 3.94 & 2.19 & 0.56 \\
\textit{Temporal setting} & \textit{The Cosmic Scale} & 3 & 3.94 & 2.19 & 0.56 \\
\textit{Tone \& Mood} & \textit{Authorial (Chimamanda Adichie)} & 3 & 3.94 & 2.19 & 0.56 \\
\textit{Tone \& Mood} & \textit{External} & 11 & 3.61 & 8.03 & 2.22 \\
\bottomrule
\end{tabularx}
\end{table}
\vspace{-1em}
Top-5 Over- or Under-selected axes by model (claude, gemini, gpt4.1). Support = pooled count of significant constraints mapped to the axis (summed across instruction types). Enrich = Share / Global, where Share = axis share within direction and Global = pooled baseline share within the same direction (computed over significant constraints, pooled across instruction types and models). Values rounded.

\subsection{Top-5 Over- or Under-selected Axes by Model (gpt5, o4mini, qwen)}\label{app:axis_top5_by_model_gpt5_o4mini_qwen}
\begin{table}[H]
\vspace{-2em}
\centering
\small
\setlength{\tabcolsep}{4pt}
\renewcommand{\arraystretch}{1.2}
\label{app:axis_top5_by_model_b}
\begin{tabularx}{\linewidth}{l X r r r r}
\toprule
\textbf{Category} & \textbf{Axis} & \textbf{Support} & \textbf{Enrich ($\times$)} & \textbf{Share (\%)} & \textbf{Global (\%)} \\
\midrule
\multicolumn{6}{l}{\textbf{gpt5 — Over-selected}}\\
\textit{Macro spatial setting} & \textit{Urban Built Environments} & 3 & 2.82 & 2.21 & 0.78 \\
\textit{Micro spatial setting} & \textit{Transit Hubs} & 3 & 2.82 & 2.21 & 0.78 \\
\textit{Relational Identity} & \textit{Defensive} & 3 & 2.82 & 2.21 & 0.78 \\
\textit{Reorientation} & \textit{Connected (E1)} & 3 & 2.82 & 2.21 & 0.78 \\
\textit{Reorientation} & \textit{Connected (E4)} & 3 & 2.82 & 2.21 & 0.78 \\
\addlinespace[2pt]
\multicolumn{6}{l}{\textbf{gpt5 — Under-selected}}\\
\textit{Embodied Difference} & \textit{Accepted} & 4 & 1.70 & 3.77 & 2.22 \\
\textit{Disruption} & \textit{Natural} & 9 & 1.58 & 2.63 & 1.67 \\
\textit{Motive} & \textit{Constructive} & 9 & 1.58 & 2.63 & 1.67 \\
\textit{Motive} & \textit{Focused} & 9 & 1.58 & 2.63 & 1.67 \\
\textit{Cultural Identity} & \textit{Legible} & 12 & 1.58 & 3.51 & 2.22 \\
\midrule
\multicolumn{6}{l}{\textbf{o4mini — Over-selected}}\\
\textit{Disruption} & \textit{Human} & 4 & 2.88 & 4.49 & 1.56 \\
\textit{Socio-political order} & \textit{Centralized} & 9 & 2.82 & 6.62 & 2.34 \\
\textit{Embodied Difference} & \textit{Race-Marked} & 6 & 2.82 & 4.41 & 1.56 \\
\textit{Temporal setting} & \textit{Realistic} & 6 & 2.82 & 4.41 & 1.56 \\
\textit{Macro spatial setting} & \textit{Urban Built Environments} & 3 & 2.82 & 2.21 & 0.78 \\
\addlinespace[2pt]
\multicolumn{6}{l}{\textbf{o4mini — Under-selected}}\\
\textit{Motive} & \textit{Constructive} & 9 & 1.83 & 3.05 & 1.67 \\
\textit{Motive} & \textit{Focused} & 9 & 1.83 & 3.05 & 1.67 \\
\textit{Cultural Identity} & \textit{Legible} & 12 & 1.83 & 4.07 & 2.22 \\
\textit{Epistemological Transformation} & \textit{Gradual} & 12 & 1.83 & 4.07 & 2.22 \\
\textit{Epistemological Transformation} & \textit{Internal} & 12 & 1.83 & 4.07 & 2.22 \\
\midrule
\multicolumn{6}{l}{\textbf{qwen — Over-selected}}\\
\textit{Motive} & \textit{Constructive} & 9 & 3.59 & 8.41 & 2.34 \\
\textit{Write like X} & \textit{Male} & 6 & 3.59 & 5.61 & 1.56 \\
\textit{Write like X} & \textit{Speculative} & 6 & 3.59 & 5.61 & 1.56 \\
\textit{Embodied Difference} & \textit{Disability-Marked} & 3 & 3.59 & 2.80 & 0.78 \\
\textit{Reorientation} & \textit{Connected (E4)} & 3 & 3.59 & 2.80 & 0.78 \\
\addlinespace[2pt]
\multicolumn{6}{l}{\textbf{qwen — Under-selected}}\\
\textit{Temporal Structure} & \textit{Future} & 6 & 4.62 & 5.13 & 1.11 \\
\textit{Reorientation} & \textit{Connected (E8)} & 3 & 4.62 & 2.56 & 0.56 \\
\textit{Temporal setting} & \textit{Age of Origins} & 3 & 4.62 & 2.56 & 0.56 \\
\textit{Tone \& Mood} & \textit{Authorial (Chimamanda Adichie)} & 3 & 4.62 & 2.56 & 0.56 \\
\textit{Write like X} & \textit{East Asian} & 5 & 3.85 & 4.27 & 1.11 \\
\bottomrule
\end{tabularx}
\end{table}
\vspace{-2.5em}
Top-5 Over- or Under-selected axes by model (gpt5, o4mini, qwen). Support = pooled count of significant constraints mapped to the axis (summed across instruction types). Enrich = Share / Global, where Share = axis share within direction and Global = pooled baseline share within the same direction (computed over significant constraints, pooled across instruction types and models). Values rounded.
\vspace{13em}

\section{Better-story and Capability User Prompts}\label{app:userprompt-variants}
\vspace{-1.4em}

\begin{table}[H]
\centering
\footnotesize
\begin{tabularx}{\linewidth}{
  >{\raggedright\arraybackslash}m{0.97\linewidth}
}
\toprule
\multicolumn{1}{c}{\fontsize{10pt}{10pt} \textbf{Experiment 2--2 (Better-story)}} \\
\midrule
\begin{minipage}[t]{\linewidth}\raggedright
\setlength{\baselineskip}{1.2\baselineskip}
As you plan to write a story, identify the specific constraints that would contribute most to high overall narrative quality in a single fictional narrative.
\medskip

\textbf{Task:}

\noindent - You will be given a list of 200 possible narrative constraints.\\
\noindent - Read through all 200 constraints carefully.\\
\noindent - Select exactly 20 constraints that would contribute most to high overall narrative quality in writing a fictional narrative.\\
\noindent - For each selected constraint, explain your reason for choosing it.\\
\noindent - After explaining your individual selections, assess the dynamics among your chosen constraints by explicitly identifying which specific constraints enhance each other and which might interfere with one another. Based on these interactions, evaluate the overall compatibility of your constraint combination and whether it would strengthen or weaken the resulting narrative when applied together in writing.\\
\noindent - There are no restrictions on the length or style of your explanations. Feel free to elaborate as much or as little as you wish.\\
\noindent - You do not need to mention constraints you are not selecting unless you wish to explain why you excluded them.\\
\noindent - List your selections using the specified output format for easy parsing.
\medskip

\textbf{Output Format:}

\noindent - Select exactly 20 constraints. The order in which you list them does not matter.\\
\noindent - For each, write only the selected constraint as a JSON object, then your reason in the "reason" field.\\
\noindent - Each constraint and its reason must appear as a separate element in a single JSON array containing all elements.\\
\noindent - After listing all selected constraints, include only one paragraph that explains the overall compatibility among all your chosen constraints as a JSON object in the form \{\{"compatibility": "[your explanation]"\}\}, and place it at the end of the array.
\medskip

\textbf{Example Output:} \{example\_lines\_2\_2\}\\
\textbf{Constraint List:} \{constraints\}
\end{minipage}
\\
\bottomrule
\end{tabularx}
\end{table}

\begin{table}[H]
\centering
\footnotesize
\begin{tabularx}{\linewidth}{
  >{\raggedright\arraybackslash}m{0.97\linewidth}
}
\toprule
\multicolumn{1}{c}{\fontsize{10pt}{10pt} \textbf{Experiment 2--2 (Capability)}} \\
\midrule
\begin{minipage}[t]{\linewidth}\raggedright
\setlength{\baselineskip}{1.2\baselineskip}
As you plan to write a story, identify the specific constraints that would contribute most to ease of execution in a single fictional narrative.
\medskip

\textbf{Task:}

\noindent - You will be given a list of 200 possible narrative constraints.\\
\noindent - Read through all 200 constraints carefully.\\
\noindent - Select exactly 20 constraints that would contribute most to ease of execution in writing a fictional narrative.\\
\noindent - For each selected constraint, explain your reason for choosing it.\\
\noindent - After explaining your individual selections, assess the dynamics among your chosen constraints by explicitly identifying which specific constraints enhance each other and which might interfere with one another. Based on these interactions, evaluate the overall compatibility of your constraint combination and whether it would strengthen or weaken the resulting narrative when applied together in writing.\\
\noindent - There are no restrictions on the length or style of your explanations. Feel free to elaborate as much or as little as you wish.\\
\noindent - You do not need to mention constraints you are not selecting unless you wish to explain why you excluded them.\\
\noindent - List your selections using the specified output format for easy parsing.
\medskip

\textbf{Output Format:}

\noindent - Select exactly 20 constraints. The order in which you list them does not matter.\\
\noindent - For each, write only the selected constraint as a JSON object, then your reason in the "reason" field.\\
\noindent - Each constraint and its reason must appear as a separate element in a single JSON array containing all elements.\\
\noindent - After listing all selected constraints, include only one paragraph that explains the overall compatibility among all your chosen constraints as a JSON object in the form \{\{"compatibility": "[your explanation]"\}\}, and place it at the end of the array.
\medskip

\textbf{Example Output:} \{example\_lines\_2\_2\}\\
\textbf{Constraint List:} \{constraints\}
\end{minipage}
\\
\bottomrule
\end{tabularx}
\end{table}

\vspace{2em}
\section{Element-Level Selection Rate Ratios Relative to \textit{Event} (gpt4.1, Better-story and Capability User Prompts)} 
\label{app:ratio-gpt4-new}
\vspace{0em}
\begin{table}[H]
\centering
\small
\setlength{\tabcolsep}{9pt}
\renewcommand{\arraystretch}{1.15}
\begin{tabular}{lcc}
\toprule
\textbf{Element} & \textbf{RR [95\% CI]} & \textbf{$p$} \\
\midrule
\textit{Style}     & 2.54 [2.38, 2.71] & < .001 \\
\textit{Character} & 0.84 [0.78, 0.92] & < .001 \\
\textit{Setting}   & 0.74 [0.68, 0.80] & < .001 \\
\bottomrule
\end{tabular}
\end{table}
\vspace{0em}
Element-level selection rate ratios relative to \textit{Event} for gpt4.1 under two additional user-prompt variants (Better-story and Capability), applied under the same Basic instruction type (pooled across the two prompt variants), with exposure offset $\log K$. \textit{N runs} $=320$ for all elements.

\end{document}